\documentclass{article}
\usepackage{amssymb}
\usepackage{times}

\usepackage{multicol}
\usepackage{graphicx}
\usepackage{wrapfig}
\usepackage{mathptmx}
\usepackage{tabularx}
\usepackage{rotating}
\usepackage{mathtools}
\usepackage{amsmath} 
\usepackage{bm}

\usepackage{amsthm}
\usepackage{siunitx}
\usepackage{wasysym}
\usepackage[mathcal]{euscript}
\usepackage{color}
\usepackage{lipsum}
\usepackage{todonotes}
\usepackage{tabularray}
\usepackage{float}
\usepackage{subcaption}
\usepackage{booktabs}
\usepackage[ruled,vlined,linesnumbered]{algorithm2e}
\usepackage{algorithmic}
\usepackage{svg}

\usepackage{multirow}
\usepackage{bbm}

\usepackage[preprint]{corl_2025} 

\usepackage[capitalise]{cleveref}
\crefname{equation}{Eq.}{Eqs.}
\crefname{figure}{Fig.}{Figs.}
\crefname{tabular}{Tab.}{Tabs.}
\crefname{section}{Sec.}{Secs.}
\crefname{algorithm}{Alg.}{Algs.}
\crefname{appendix}{App.}{Apps.}

\newcommand{\regtext}[1]{\mathrm{\textnormal{#1}}}

\newcommand{\hypothesis}[1]{\textbf{\textit{#1}}}

\definecolor{darkpink}{rgb}{0.61, 0.33, 0.5}

\newif\ifshowcomments
\showcommentsfalse

\newcommand{\norm}[1]{\left\Vert#1\right\Vert}

\newcommand{\trans}{^\top}

\newcommand{\lbl}[1]{^{\regtext{#1}}}

\newcommand{\des}{\lbl{d}}
\newcommand{\plan}{\lbl{p}}
\newcommand{\exec}{\lbl{e}}
\newcommand{\delay}{\lbl{delay}}
\newcommand{\cond}{\lbl{c}}
\newcommand{\fast}{\lbl{fast}}
\newcommand{\slow}{\lbl{slow}}
\newcommand{\orig}{\lbl{*}}
\newcommand{\future}{\lbl{f}}
\newcommand{\obs}{\lbl{$\observation$}}
\newcommand{\act}{\lbl{$\action$}}

\newcommand\freefootnote[1]{%
  \let\thefootnote\relax%
  \footnotetext{#1}%
  \let\thefootnote\svthefootnote%
}

\newcommand{\dataset}{\mathcal{D}}
\newcommand{\controller}{\mathcal{K}}
\newcommand{\policy}{\pi}
\newcommand{\state}{x}
\newcommand{\action}{a}

\newcommand{\horizon}{H}
\newcommand{\control}{u}
\newcommand{\torque}{\tau}
\newcommand{\jacobian}{J}
\newcommand{\massmatrix}{M}
\newcommand{\gain}{K}
\newcommand{\dt}{{\delta}}
\newcommand{\dtoriginal}{{\delta\orig}}
\newcommand{\dtbound}{\dt\lbl{lb}}
\newcommand{\error}{e}
\newcommand{\errorbound}{\rho}
\newcommand{\observation}{o}
\newcommand{\critactflag}{k}
\newcommand{\speedupfactor}{c}
\newcommand{\angularfreq}{\omega}
\newcommand{\freqcutoff}{\omega_c}
\newcommand{\fouriermagnitudespectrum}{V}
\newcommand{\speedprofile}{v}
\DeclareMathOperator{\errorfunction}{error}

\DeclareMathOperator{\TPR}{TPR}

\newcommand{\noisepred}{\epsilon_\theta}

\newcommand{\weight}{w}

\newcommand{\futureactionhorizon}{\horizon\lbl{f}}

\newcommand{\conditioningactions}{\action\cond}

\newcommand{\trackerr}{e}

\title{\emph{SAIL}: Faster-than-Demonstration Execution of Imitation Learning Policies}

\author{%
  Nadun Ranawaka Arachchige\thanks{Equal contribution}, $\;$ Zhenyang Chen\footnotemark[1],$\;$ Wonsuhk Jung, \\
  \bf Woo Chul Shin, $\;$ Rohan Bansal, $\;$ Pierre Barroso, $\;$ Yu Hang He, $\;$ \\
  \bf Yingyan Celine Lin, $\;$ Benjamin Joffe, $\;$ Shreyas Kousik\thanks{Equal advising. All authors are with the Georgia Institute of Technology, Atlanta GA, USA. \\Correspondence: \texttt{nadun.ranawaka@gatech.edu}}, $\;$ Danfei Xu\footnotemark[2] \\[4pt]
  \normalsize Georgia Institute of Technology
}

\begin{document}
\maketitle

\begin{abstract}
     Offline Imitation Learning (IL) methods such as Behavior Cloning are effective at acquiring complex robotic manipulation skills. 
     However, existing IL-trained policies are confined to executing the task at the same speed as shown in demonstration data. This limits the \emph{task throughput} of a robotic system, a critical requirement for applications such as industrial automation. In this paper, we introduce and formalize the novel problem of enabling faster-than-demonstration execution of visuomotor policies and identify fundamental challenges in robot dynamics and state-action distribution shifts. We instantiate the key insights as SAIL (Speed Adaptation for Imitation Learning), a \emph{full-stack system} integrating four tightly-connected components: (1) a consistency-preserving action inference algorithm for smooth motion at high speed, (2) high-fidelity tracking of controller-invariant motion targets, (3) adaptive speed modulation that dynamically adjusts execution speed based on motion complexity, and (4) action scheduling to handle real-world system latencies. 
      Experiments on 12 tasks across simulation and two real, distinct robot platforms show that SAIL achieves up to a {4$\times$ speedup} over demonstration speed in simulation and up to {3.2$\times$ speedup} in the real world. {Additional detail is available at \url{https://nadunranawaka1.github.io/sail-policy}}

\end{abstract}




\section{Introduction}

\begin{wrapfigure}{r}{0.5\textwidth}  
  \vspace{-20pt}                        
  \centering
  \includegraphics[width=0.47\textwidth]{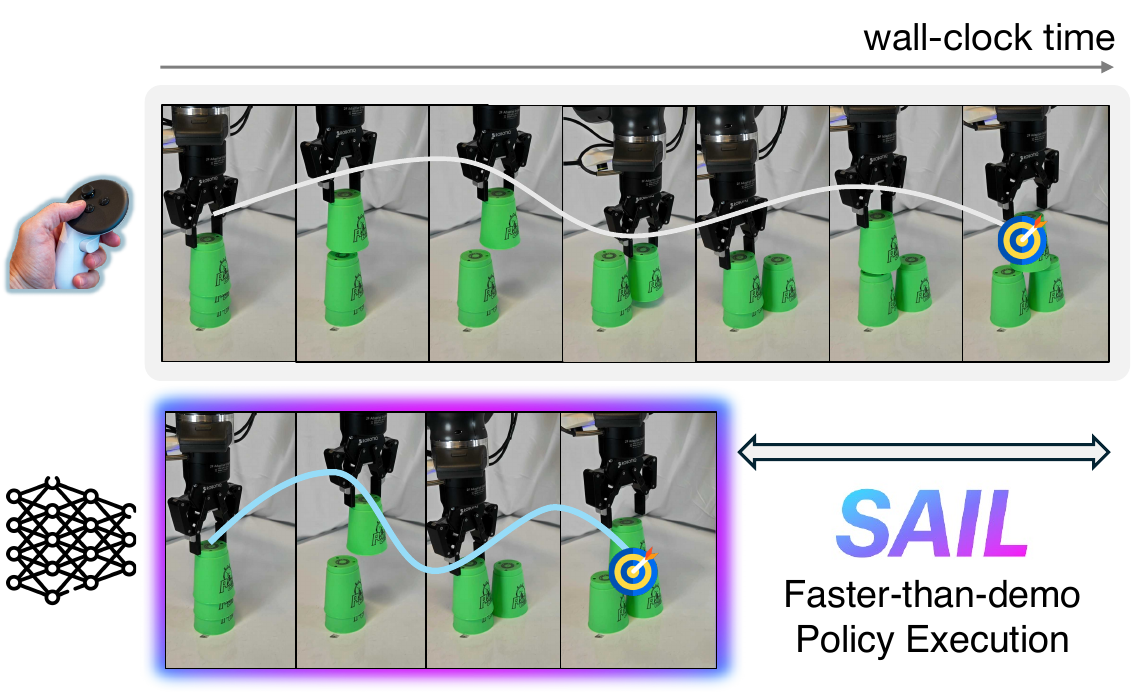}
  \caption{The goal of our system, Speed‑Adaptive Imitation Learning (SAIL), is to speed up the execution of a learned visuomotor policy such that the robot can complete manipulation tasks faster than in the original training demonstrations.}
  \label{fig:SAIL_teaser}
  \vspace{-10pt}                        
\end{wrapfigure}

Speed is essential for real-world robot learning applications. Recent offline imitation learning methods~\cite{chi2023diffusionpolicy, zhao2023learning} have excelled in complex tasks like deformable object manipulation and non-prehensile actions. However, their performance heavily depends on human demonstrations, which are typically slow, leading policies to inherit sluggish motions. Therefore, this paper addresses the question: \emph{How can we speed up the execution of learned visuomotor policies beyond demonstration speed?} A key challenge is that increased execution speeds alter robot dynamics, introducing tracking errors and dynamic effects, which in turn shifts the observation distribution and further causes the policy to deviate from its prior distribution. Moreover, practical acceleration is constrained by inference latency, sensor delays, and control bandwidth~\cite{sakaino2022imitation}, making speed-up a full-stack challenge. As a result, previous works typically assume consistent execution speeds during training and deployment.

We introduce \textbf{S}peed \textbf{A}daptation for \textbf{I}mitation \textbf{L}earning (SAIL) as a full-stack framework to tackle this problem (\cref{fig:SAIL_teaser}).
Our key insight is that successfully accelerating policy execution requires addressing both \emph{changing robot dynamics} and the resulting \emph{state-action distribution shifts}.
At the policy level, SAIL features: (1) an controller error-aware generative guidance algorithm that generates temporally-consistent action predictions at varying execution speed (\Cref{subsec:method:cfg}); and (2) adaptive speed modulation that dynamically adjusts execution speed based on task demands, slowing down for precision and accelerating otherwise (\cref{subsec:adaptive_speed_execution}).
At the system level, SAIL: (1) trains the policy model to predict controller-invariant action targets and tracks the targets with high-fidelity controller to mitigate dynamic shifts (\Cref{subsec:enhanced_controller}); and (2) adaptively schedules action execution to maintain real-time control despite sensing and computation delays (\Cref{subsec:method:action_scheduling}).

This full-stack approach allows SAIL to achieve significantly higher \emph{task throughput} across a variety of manipulation tasks.
We validate our key technical insights and design choices through experiments in simulation and demonstrate that SAIL achieves up to a $4\times$ speedup over demonstration speed while maintaining high task success rates. We also show that, on two physical robot systems with distinctive controller and dynamics, SAIL achieves up to a {$3.2\times$} speedup across 7 tasks involving challenges such as long task horizons, high precision, and bimanual coordination.
\section{Related Work} 

\noindent\textbf{Offline Imitation Learning.} Offline imitation learning (LfD) methods are common ways to program robots to learn behaviors from human demonstrations~\cite{pomerleau1988alvinn,argall2009survey,hussein2017imitation,ravichandar2020recent}. Particularly, behavior cloning (BC)~\cite{ravichandar2020recent} trains deep networks to map observations to actions from demonstration data. Recent works in BC have introduced deep generative models~\cite{zhao2023learning,chi2023diffusionpolicy} to preserve the multimodal aspects of real-world trajectories. However, as noted by prior works~\cite{zhao2023learning,liu2024bidirectional}, sampling from these learned trajectory distributions can break temporal dependencies between consecutive prediction steps. This challenge is further exacerbated under higher execution speeds. Relevant to our goal of speed-adaptive imitation are state-space frameworks like Dynamic Movement Primitives (DMPs)~\cite{1041514, 976259, DBLP:journals/corr/abs-2102-03861} and Riemannian Motion Policies (RMPs)~\cite{DBLP:journals/corr/abs-1801-02854}, which theoretically support temporal modulation of motion. Although a few approaches~\cite{bahl2020neural, schaal2003control, xie2022neural, DBLP:journals/corr/abs-2107-05627} have incorporated such techniques into end-to-end imitation learning, the field has more widely embraced simpler algorithms like Diffusion Policy~\cite{chi2023diffusionpolicy}.

\noindent\textbf{Better-than-Demonstration Imitation Learning.}
There are multiple lines of work that aim to learn better-than-expert policies~\cite{brown2019extrapolating, brown2020better, wu2019imitation}. For instance, T-REX~\cite{brown2019extrapolating} learns better-than-demonstration policies by using a ranking-based reward function to evaluate unseen policy behaviors. Others~\cite{sakaino2022imitation,saigusa2021imitation} rely on subsequent self-supervised online learning with task-specific reward design~\cite{saigusa2022imitation}, enabling a 1.1--1.3$\times$ increase in execution speed at the expense of a loss in success rate. Furthermore, these methods typically operate in an Inverse Reinforcement Learning~\cite{ng2000algorithms, abbeel2004apprenticeship, ziebart2008maximum} setting, requiring interactive learning in the environment. In contrast, our focus is on the purely offline setting, where we execute an offline-learned policy faster during runtime. 

Recent works, such as SPHINX~\cite{sundaresan2024s}, have shown modest improvements in execution time and speedup. Nevertheless, these gains are typically byproducts of their methods, whereas we explicitly target faster execution as a primary objective. 

\noindent\textbf{System Integration for Imitation Learning.}
Recent works have expanded beyond pure algorithmic innovations to develop full-stack systems for imitation learning~\cite{zhao2023learning,fu2024mobile,cheng2024tv,iyer2024open,chi2024universal}, 

with systems such as Mobile-ALOHA~\cite{fu2024mobile} and UMI~\cite{chi2024universal} showing tremendous value in full-stack optimization of low-level robot controllers with learning algorithms.
However, none of these works consider the problem of deliberately varying the execution speed between demonstration and policy execution. Our core contribution lies in identifying key challenges for this new problem setting and proposing a system that enables faster-than-demonstration policy execution.
\section{Preliminaries, Challenges, and Problem Statement}
\label{sec:problem}

\textbf{Policy and Controller Hierarchy.} We consider a robot control system structured in two levels: (1) a high-level neural network policy $\policy(\state, \observation)$ generating an action command $\action$ based on the current robot state $\state$ and sensory observation $\observation$, and (2) a low-level robot controller $\controller$ translating $\action$ into robot joint torques $\control$.
To make our method generally applicable, we follow a common setup~\cite{chi2023diffusionpolicy, chi2024universal,liu2024rdt} and assume that $\policy$ outputs action trajectories $\action_t = [\state\des_t, \state\des_{t+1}, \cdots, \state\des_{t+\horizon-1}]$, where $H$ is the prediction horizon and each $\state\des_i$ includes the desired SE(3) end-effector pose and gripper command.

The controller $\controller$ tracks this trajectory, calculating the torque $\control_t$ for each desired pose $\state\des_t$ given the current state $\state_t$ and a fixed time interval $\dtoriginal$ between reference configurations: $\control_t = \controller(\state\des_t, \state_t, \dtoriginal)$. The controller $\controller$ typically operates at a significantly higher frequency than the policy inference rate, executing multiple control steps to track each $\state\des_t$. Policies are usually deployed in a receding-horizon fashion, executing a portion of the trajectory ($\horizon\exec < \horizon$ steps) before replanning.

\textbf{Offline Imitation Learning.} We focus on offline imitation learning, where the policy $\policy$ is learned from a static dataset $\dataset = \{(\observation_t, \state_t, \state\des_t)\}_{t=1}^{n}$ collected via teleoperation. Each datum contains the sensory observation $\observation_t$, robot state $\state_t$, and the desired configuration $\state\des_t$ commanded by the teleoperator, sampled at a fixed time interval $\dtoriginal$.
The action supervision $\action_t$ for training is formed by extracting sequences of future desired configurations $[\state\des_t, \cdots, \state\des_{t+\horizon-1}]$ from the demonstration.
In this work, we focus on Diffusion Policy (DP)~\cite{chi2023diffusionpolicy} as our representative high-performance policy model. 
DP generates multi-step action trajectories via iterative denoising based on the current state and observation $(\observation_t, \state_t)$, following $\action_n = \action_{n+1} - \gamma\epsilon_\theta(\observation_t, \state_t, \action_{n+1}, n) + \mathcal{N}(0,\sigma_n^2\mathbf{I})$, where $\epsilon_\theta$ is the learned denoising network. This action chunking~\cite{chi2023diffusionpolicy,zhao2023learning,liu2024rdt} capability is advantageous for handling inference latency. However, finite training data and the probabilistic nature of the policy can lead to temporal inconsistencies between consecutively predicted trajectories~\cite{liu2024bidirectional}, potentially causing jerky motion, especially at high speeds. Furthermore, network inference introduces latency that must be managed for real-time, high-speed control (\Cref{subsec:method:action_scheduling}). We note that all but one component in our system can apply to other generative policy models, such as ACT~\cite{zhao2023learning}, which we show in \Cref{appendix:subsec:ablation}.

\textbf{Challenges and Problem Statement.} Executing learned policies faster than demonstrated introduces critical challenges beyond standard imitation learning. 
Faster execution drastically exacerbates \emph{distribution shift}, as altered system dynamics and controller errors push the policy into unfamiliar Out-of-Distribution states where compounding errors~\cite{ross2011reduction} lead to failure.
Simultaneously, achieving speedup strains \emph{execution fidelity}: low-level controllers may struggle to track fast trajectories accurately, the assumption of consistent controller behavior between demonstration and execution breaks down, and tasks inherently require slower speeds during high-precision phases. Finally, real-world \emph{system latencies} impose hard physical limits on control loop frequency, fundamentally constraining maximum achievable speed.
\emph{Therefore, our core research problem is}: Given a policy $\policy$ trained with time interval $\dt\orig$, how can we execute it using a faster, time-varying time interval $\dt_t = \speedupfactor_t \dt\orig$ (with speedup factor $\speedupfactor_t \leq 1$) to significantly increase successful task throughput while overcoming the intertwined challenges of distribution shift, execution fidelity, and latency?

\section{Speed Adaptation for Imitation Learning (SAIL)}
\label{section:method}

\subsection{Consistent Action Prediction via Error-Adaptive Guidance}
\label{subsec:method:cfg}
\begin{wrapfigure}{r}{0.5\textwidth}   
  \centering
  \vspace{-10pt} 
  \includegraphics[width=0.5\textwidth]{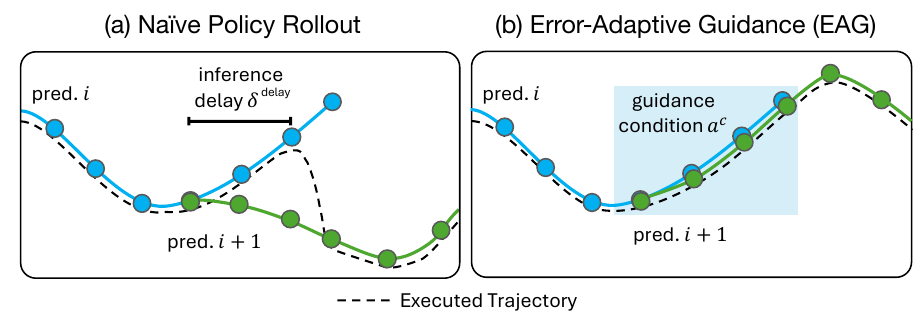}
  \caption{\textbf{Divergence during receding horizon execution.}
  We found that the naive policy rollout occasionally produces inconsistent predictions between planning iterations, as shown in (a). For example, the blue and green trajectories are two consecutive trajectories that diverge in path. This can cause a jerky executed trajectory (black dashed line) during receding horizon control. EAG addresses this problem by enforcing consistent consecutive predictions via conditional guidance $a^c$, resulting in smoother execution.}
  \vspace{-8pt} \label{fig:diverging_trajectory_predictions}
\end{wrapfigure}

Executing policies significantly faster than demonstrated fundamentally alters controller dynamics.
This leads to increased tracking error, pushing the robot state $(\observation_t, \state_t)$ Out-of-Distribution (OOD) compared to its training experience captured in dataset $\dataset$. This OOD shift poses a critical challenge during receding horizon execution with visuomotor policies. When faced with OOD inputs resulting from tracking errors in the previous execution step, the policy $\policy$ can produce subsequent action predictions $\hat{\action}_{0:\horizon}$ that are inconsistent with the just-executed actions $\action_{\horizon\exec:\horizon\exec+\futureactionhorizon}$.
As shown in \Cref{fig:diverging_trajectory_predictions}, this prediction \textbf{divergence} between planning steps manifests as jerky or unstable robot motion.

To mitigate this, one might consider enforcing temporal smoothness using Classifier-Free Guidance (CFG)~\cite{ho2022classifier}.
CFG encourages consistency by conditioning the prediction of the next action sequence $\hat{\action}_{0:\horizon}$ on the tail of the previously planned sequence $\action\cond = \action_{\horizon\exec:\horizon\exec+\futureactionhorizon}$.
This is achieved by blending the conditional score estimate $\noisepred(\action\future, \action\cond | \observation_t, \state_t)$ with the unconditional one $\noisepred(\action\future, \emptyset | \observation_t, \state_t)$ using a guidance weight $\weight$ as
\begin{equation} 
    \noisepred\lbl{guided}(\action\future, \action\cond | \observation_t, \state_t) = 
        \noisepred(\action\future, \emptyset | \observation_t, \state_t) + 
        \weight(\noisepred(\action\future, \action\cond | \observation_t, \state_t) - 
        \noisepred(\action\future, \emptyset | \observation_t, \state_t)).
\end{equation}

However, naively applying CFG guidance ($\weight > 0$) during high-speed execution can be detrimental. If significant tracking error occurred while executing $\action\cond$, the input $(\state_t, \action\cond)$ to the conditional model $\noisepred(\cdot, \action\cond | \observation_t, \state_t)$ is effectively OOD. Conditioning strongly on this unreliable information provides misleading guidance, potentially further exacerbating divergence. This challenge highlights the limitations of approaches that either perform smoothing \emph{post-hoc} after action generation~\cite{liu2024bidirectional,zhao2023learning} or implicitly assume consistent execution dynamics~\cite{chi2023diffusionpolicy,hoeg2024streaming} without directly addressing the OOD inputs caused by controller shifts during the generation process itself.

\begin{figure*}[t]
    \centering
    \includegraphics[width=\textwidth]{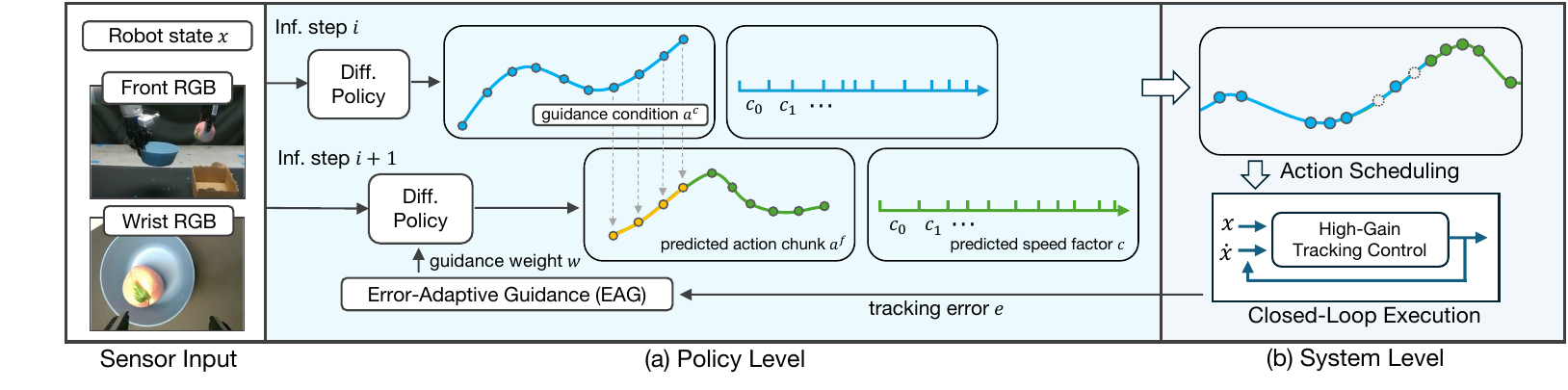} 
    \caption{\textbf{System Overview.} SAIL operates at two levels: (a) \textbf{Policy Level}: Given raw sensor input, the policy generates (1) temporally-consistent action predictions through error-adaptive guidance (EAG) and (2) time-varying speedup factor. (b) \textbf{System Level}: The predicted actions are scheduled for execution while accounting for sensing-inference delays, with outdated actions being discarded. The actions are tracked with a high-fidelity controller at the specified time parametrization.}
    \vspace{-10pt}
    \label{fig:system}
\end{figure*}

Therefore, we propose \textbf{Error-Adaptive Guidance (EAG)}. We recognize that the utility of conditioning via CFG depends on the quality of the conditioning signal, which degrades with increasing tracking error.
We use the current end-effector tracking error $\error = \errorfunction(\state\des, \state\lbl{current})$ as an efficient proxy for this potential OOD state (validated in Appendix \ref{appendix:subsec:track_err}). Our algorithm dynamically adjusts the CFG guidance weight $\weight$: if tracking is accurate ($\error \le \errorbound$), we apply standard CFG guidance ($\weight > 0$) assuming reliable conditioning. Conversely, if the error is large ($\error > \errorbound$), indicating the conditioning input $\action\cond$ is likely unreliable due to the OOD state, we disable guidance ($\weight = 0$) and rely solely on the unconditional prediction $\noisepred(\action\future, \emptyset | \observation_t, \state_t)$.  This adaptive approach allows SAIL to enforce temporal consistency when tracking is accurate, while preserving robustness by avoiding potentially incorrect guidance caused by controller shift. We illustrate how EAG is integrated with the overall system in \cref{fig:system} and detail its implementation in \cref{alg:EAG}.

\subsection{Reducing Controller Shift via Controller-invariant Action Target}
\label{subsec:enhanced_controller}
\label{subsec:method:control}

\begin{wrapfigure}{r}{0.5\textwidth}  
  \vspace{-10pt}                        
  \centering
  \includegraphics[width=0.47\textwidth]{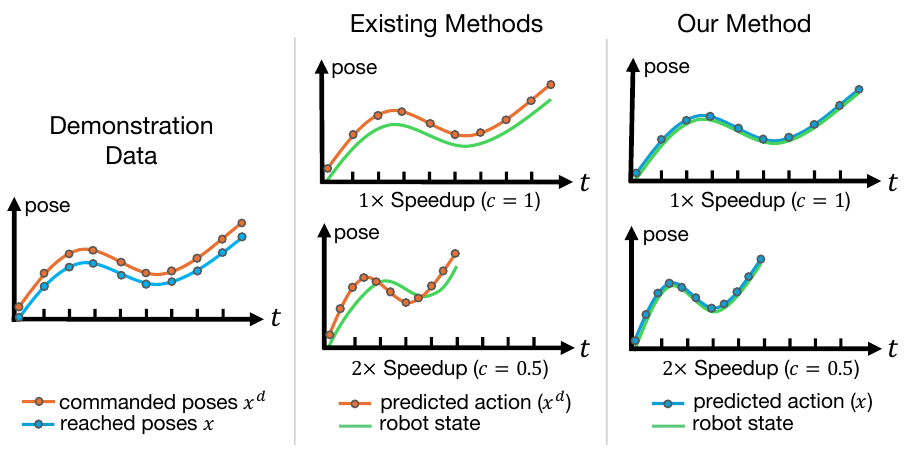}
  \caption{\textbf{Commanded vs Reached Pose.}
    (Left) Teleoperator commands  $\state\des$ to the robot, and the robot reaches poses $\state$. (Middle) Most policies are trained to predict $\state\des$ and suffer from error profile change during speeding up execution. (Right)
    We minimize this shift by training policies to predict the reached poses $\state$ and track these with a high-fidelity controller.}
  \label{fig:controller}
  \vspace{-10pt}                        
\end{wrapfigure}

While \Cref{subsec:method:cfg} adaptively handles Out-of-Distribution (OOD) states algorithmically, a primary source of this distribution shift during high-speed execution is the changing behavior of the low-level controller. Standard Imitation Learning often trains policies $\policy$ to predict the \textbf{commanded} poses $\state\des$ from teleoperation (collected at interval $\dt\orig$, often with a low-gain controller to ensure smooth user experience) and uses the same teleoperation controller $\controller\lbl{teleop}$ for execution. However, running $\controller\lbl{teleop}$ at a faster interval $\delta_t < \dt\orig$ alters its dynamics and tracking error profile, providing OOD state inputs to the policy and hindering performance (\Cref{fig:controller}, Middle).

To mitigate this at the system level, we first change \emph{what} the policy predicts. Instead of the commanded pose $\state\des$, we train $\policy$ to predict the \textbf{reached pose} $\state$ directly from the demonstration data (\Cref{fig:controller}, Right). The reached pose represents the robot's actual trajectory and is inherently achievable. Crucially, this target is largely \emph{invariant} to the specific dynamics of the potentially compliant or noisy $\controller\lbl{teleop}$ used during data collection, providing a more stable prediction target for the policy across different execution speeds.
Complementing this, we also change \emph{how} the predicted action is executed. During high-speed deployment, we replace the original $\controller\lbl{teleop}$ with a dedicated, \emph{high-fidelity tracking controller} $\controller\exec$ optimized for fast and accurate motion. This controller takes the policy's predicted reached poses $\state_t$ as targets and generates torques $\control_t = \controller\exec(\state_t, \state\lbl{current}, \delta_t)$ to follow them closely, even at the reduced interval $\dt_t$. Using a high-performance controller (like our high-gain OSC detailed in \Cref{appendix:sec:real_robot_setup} or potentially MPC) to track the invariant reached pose target effectively decouples policy execution from the variable dynamics of the teleoperation controller,  reducing controller-induced distribution shift.

\subsection{Adaptive-Speed Policy Execution}
\label{subsec:adaptive_speed_execution}

Executing all parts of a manipulation task at maximum speed can compromise success, especially during precise interactions like grasping or alignment.
SAIL automatically identifies such \emph{critical actions} and dynamically adjusts the instantaneous speedup factor $\speedupfactor_t$ based on the real-time context, slowing down during critical actions and speeding up otherwise.

We identify critical actions using two complementary methods based on analyzing demonstration data and runtime predictions.
(1) \emph{Motion Complexity Analysis}: Inspired by~\cite{shi2023waypoint}, we perform offline analysis (e.g., DBSCAN clustering~\cite{mester1996dbscan} on demonstration waypoints) to flag segments with high geometric complexity.
We then train the policy $\policy$ to predict, alongside its action, whether the current action corresponds to a phase that requires complex and careful motion.
(2) \emph{Gripper Event Detection}: A simple heuristic to identify likely interaction phases at runtime is detecting changes (opening/closing) in the predicted gripper state within the policy's output action sequence $\action_{t:t+\horizon}$.
Both of these methods yield a binary critical action flag $\critactflag_t \in \{0,1\}$ for the current timestep $t$, where $\critactflag_t = 1$ indicates a critical action.
We modulate the speedup factor between preset slow ($\speedupfactor\slow$) and fast ($\speedupfactor\fast$) values:
$\speedupfactor_t = \critactflag_t \cdot \speedupfactor\slow + (1-\critactflag_t) \cdot \speedupfactor\fast$.
This allows SAIL to automatically slow down for precision and speed up during simpler motions like reaching.
We include more details in \Cref{subsec:appendix:adaptive_speed}.

\subsection{Maintaining Real-Time Control at High Speed Under System Latency} 
\label{subsec:method:action_scheduling}

\textbf{Action scheduling.} While adaptive modulation determines the \emph{desired} execution speed $\speedupfactor_t$, achieving stable control at high average speeds requires explicitly handling inherent system latencies~\cite{chi2024universal}. There exists an irreducible sensing-to-action delay $\dt\delay$ between requesting sensor data (at $t\obs$) and receiving the computed action sequence from the policy (at $t\act$). To maintain continuous robot motion and prevent pauses, SAIL executes actions planned from the \emph{previous} policy inference step during this $\dt\delay$ interval. Once the new action sequence $\hat{\action}_{0:H}$ arrives at $t^\action$, the system schedules its execution using the adaptively determined interval $\delta_t = \speedupfactor_t \dt\orig$, transitioning from the old plan by discarding any actions from the previous plan scheduled after $t\act$. This process is illustrated in \Cref{fig:action_scheduling}.

\textbf{Upper bounding speedup.} Crucially, the latency $\dt\delay$ imposes a fundamental physical limit on the maximum achievable average speedup. To prevent ``action exhaustion''---running out of commands before the next inference cycle completes---the actual execution interval $\delta_t$ must be lower-bounded. This lower bound, $\delta\lbl{lb} > \dt\delay / (\horizon\plan - \horizon\cond)$ (where $\horizon\plan$ is prediction horizon and $\horizon\cond$ is conditioning length, derived in \Cref{appendix:derivation_lower_bound}), represents the minimum time required per step to sustain continuous operation. Therefore, the adaptively chosen desired interval $\speedupfactor_t \dt\orig$ must respect this physical limit. The final execution interval used by the controller at time $t$ is $\delta_t = \max(\speedupfactor_t \cdot \dt\orig, \delta\lbl{lb})$. This ensures SAIL dynamically varies speed according to $\speedupfactor_t$, but never exceeds the system's physical capacity due to latency, even at high speeds.
\section{Evaluations}\label{sec:experiments}

We test the following hypotheses:
\hypothesis{H1}: Error-Aware Guidance (EAG) generates temporally-consistent actions that improve policy performance at high speed,
\hypothesis{H2}: All components of SAIL are critical for enabling faster-than-demo execution while keeping a high success rate, and
\hypothesis{H3}: SAIL is generally deployable to physical robots on realistic tasks.

\textbf{Metrics.}
 Our primary evaluation metric is throughput-with-regret (TPR$\uparrow$), which rewards faster successes while penalizing all failures equally, thus reflecting a policy's capability to maintain success rates at higher execution speeds.
We further consider two suites of metrics. The first suite focuses on task performance and efficiency, which includes task success rate (SR$\uparrow$), TPR, average time for successful rollouts (ATR$\downarrow$), speedup-over-demo (SOD$\uparrow$).
The second suite characterizes the generated motion trajectories (e.g., smoothness), including consistency (CON$\downarrow$), spectral arc length~\cite{balasubramanian2015smoothness} (SPARC$\uparrow$), log dimensionless jerk~\cite{balasubramanian2015smoothness} (LDLJ$\downarrow$), and weighted Euclidean distance~\cite{liu2024bidirectional} (WED$\downarrow$).
Detailed metric descriptions are in \Cref{ssec:appendix:metrics}.

\subsection{Simulation Evaluation: SAIL Achieves High Task Throughput}
\label{sec:exp:h4-5}

\textbf{Setup.}
We evaluate SAIL and a variety of baselines on standard manipulation benchmarks from RoboMimic~\cite{mandlekarmatters} and MimicGen~\cite{mandlekar2023mimicgen}: Lift, Can, Square, Stack, and Mug Cleanup. 
More details on the experiment setup are presented in \Cref{appendix:sim_exp_details}.
%
Since few imitation learning methods directly address execution speedup, we design several baselines for comparison, including AWE~\cite{shi2023waypoint} which achieves speedup as a side effect of its waypoint-based approach. Our baselines are: (1) \textbf{DP}~\cite{chi2023diffusionpolicy}: Executes actions at the original demonstration speed, serving as our primary baseline. (2) \textbf{DP-Fast}: Executes Diffusion Policy actions at an accelerated fixed frequency using a low-gain controller, representing the naive approach to speedup. (3) \textbf{Aggregated Actions}: Operates in delta Cartesian Space by aggregating consecutive actions in similar directions (detailed in \Cref{ssec:appendix:baseline}). (4) \textbf{AWE}~\cite{shi2023waypoint}: Uses automatically extracted waypoints to generate absolute action labels. (5) \textbf{BID-Fast}: the same as \textbf{DP-Fast} but using BID~\cite{liu2024bidirectional} to enforce consistency between consecutive predictions during receding horizon rollouts. See the architectures and parameter details in \Cref{appendix:hyperparameters}.

\begin{figure}[ht] 
    \centering
    \includegraphics[width=\columnwidth]{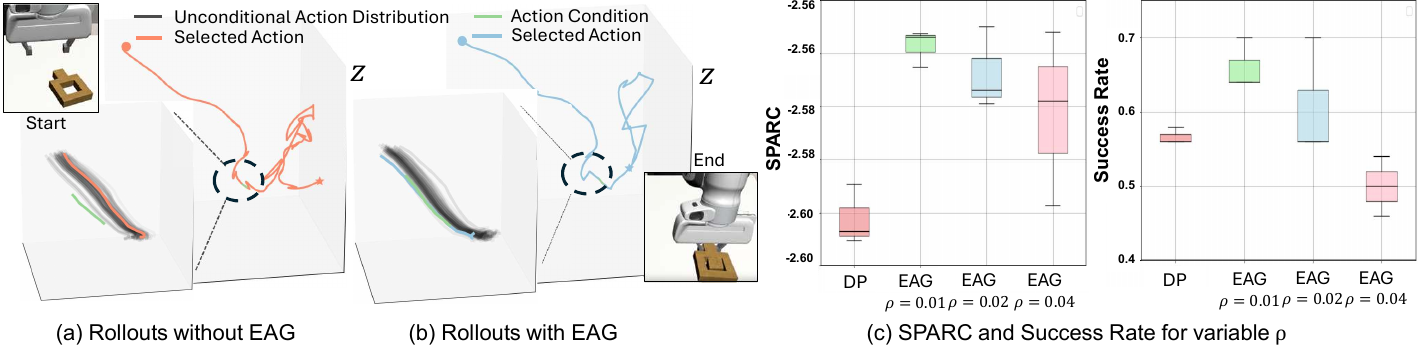}
    \caption{We show that EAG generates temporally-consistent motion, comparing sample {rollouts without EAG} in (a) and {rollouts with EAG} in (b). We further illustrate how the success rate and smoothness of trajectories are affected by the error threshold $\rho$ in (c).}
    \label{fig:EAG_result}                       
    \vspace{-1pt} 
\end{figure}

\textbf{EAG generates temporally-consistent motion at high speeds (\hypothesis{H1}).}

To assess the effectiveness of EAG, we evaluate both trajectory smoothness and task performance. Qualitative motion results and quantitative metric (SPARC) in \Cref{fig:EAG_result} shows that EAG significantly enhances motion consistency, which in turn improves task success rates during accelerated execution (SR). This is further confirmed by the ablation study \textbf{SAIL(-C)} in \Cref{tab:sim_ablation_full}, where removing consistency-preserving trajectory generation reduces performance. Additionally, we observe that consistent guidance is most effective when the conditioned future action falls within the unconditional action distribution. As execution speed increases, misalignment between observations and conditioned actions becomes more pronounced, correlating strongly with increased tracking error. Detailed results presented in \Cref{appendix:subsec:CFG_experiments} confirm that adaptive tracking error cutoff is crucial for effective conditional guidance.

\begin{wrapfigure}{r}{0.5\textwidth}  
  \vspace{-10pt}                        
  \centering
  \includegraphics[width=0.5\textwidth]{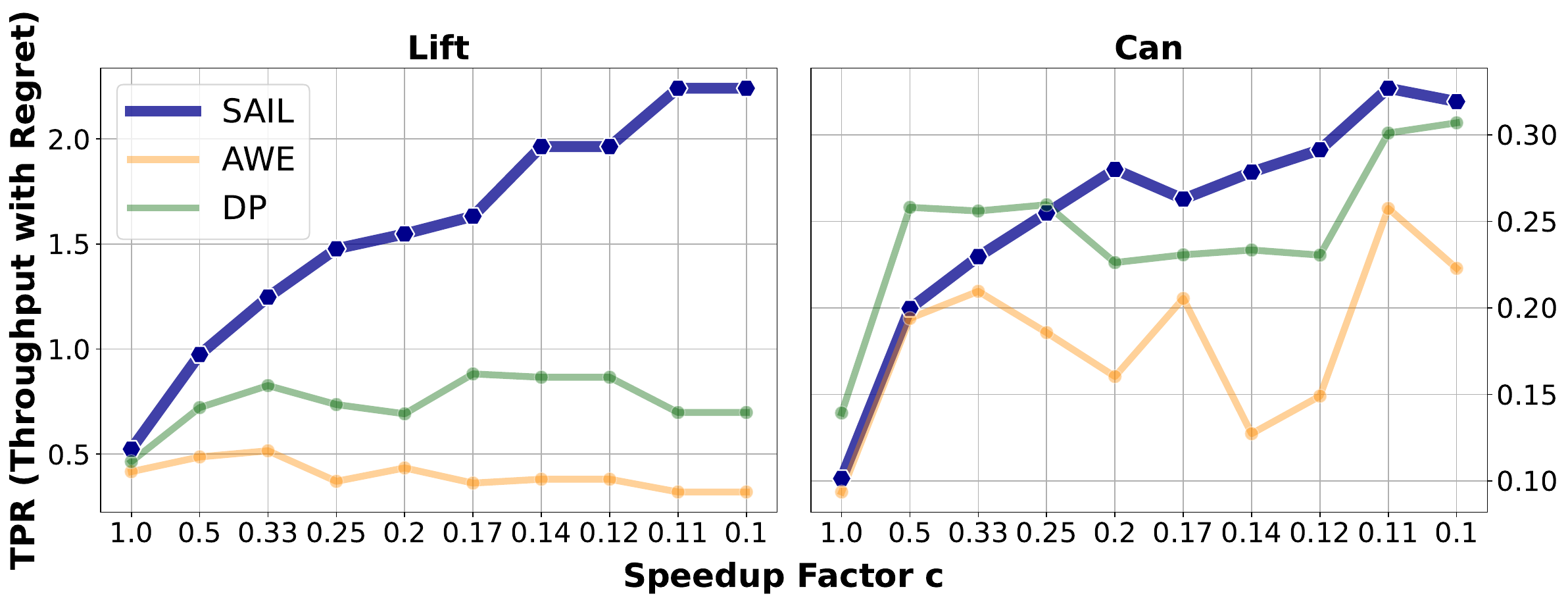}
  \caption{
    \textbf{TPR vs. Speedup Factor on Can and Lift Tasks.}
    We show that as the speedup factor increases, SAIL's throughput-with-regret increases more than the AWE and DP baselines.
    In other words, SAIL is able to accumulate more task successes more quickly while limiting task failures.}
  \label{fig:c_vs_TPR}
  \vspace{-10pt}                        
\end{wrapfigure}

\textbf{SAIL achieves much higher throughput than baselines (\hypothesis{H1}).}
As seen in \Cref{tab:simeval}, SAIL can achieve up to $3\times$ throughput of baselines such as DP~\cite{chi2023diffusionpolicy} for the \textbf{Can} and \textbf{Stack} tasks, without sacrificing success rate. We attribute this to the combination of components, including the high-gain controller, tracking reached poses, and adaptive speed modulation.
Moreover, we conducted a thorough study of how the throughput (TPR) changes as we vary $\speedupfactor_t$ in \Cref{fig:c_vs_TPR}.
We observe that SAIL is able to smoothly improve the TPR as $\speedupfactor_t$ increases, with up to $\speedupfactor_t = 0.1$ ($10\times$ speedup) showing its robustness at least in an ideal simulated environment.

\textbf{Component Ablation (\hypothesis{H2}).}
We present the conclusion of ablation studies evaluating key components in SAIL, with full results in the Appendix. First, high-gain controllers, while necessary for fast tracking, are sensitive to reference trajectory noise. As shown in \Cref{fig:gain_vs_noise} in \cref{subsec:exp2}, noisy references drastically reduce success rates for high-gain control, highlighting the need for smooth references generated by mechanisms like EAG. Second, imitating reached poses is critical for high-speed execution. Substituting commanded poses significantly degrades performance, particularly when replaying demonstrations faster or with higher gains (details in \Cref{subsec:exp1}). This degradation is quantified in \Cref{tab:sim_ablation_full}: using commanded poses instead of reached poses results in a $55\%$ drop in success rate for \textbf{Square} and reduces the average TPR by $0.08$ across tasks compared to full SAIL. Finally, adaptive speed modulation (AS) proves vital, especially for precision tasks. Comparing SAIL with \textbf{SAIL(-AS)} (SAIL without adaptive speed) in \Cref{tab:sim_ablation_full}, we observe significantly lower success rates without AS, particularly on \textbf{Square} and \textbf{Mug} (degradations up to $31\%$). This underscores the importance of dynamically adjusting speed, especially for high-precision scenarios like \textbf{Square}.

\begin{table}[ht]
\centering
\caption{Results of evaluation in simulation (Robomimic~\cite{mandlekarmatters} and MimicGen~\cite{mandlekar2023mimicgen})}
\label{tab:simeval}

\scriptsize                    
\begin{tblr}{
  width  = \linewidth,         
  colsep = 0.5pt,                
  rowsep = 1pt,
  colspec = {Q[l] | *{5}{XXXX}},   
  row{1} = {font=\bfseries},        
  hline{1,3,9} = {-}{},             
  vline{6,10,14,18} = {1-8}{}, 
  cell{1}{2}  = {c=4}{c},
  cell{1}{6}  = {c=4}{c},
  cell{1}{10} = {c=4}{c},
  cell{1}{14} = {c=4}{c},
  cell{1}{18} = {c=4}{c},
}

& Lift & & & &
  Can  & & & &
  Square & & & &
  Stack  & & & &
  Mug \\

Method &
SR$\uparrow$ & TPR$\uparrow$ & ATR$\downarrow$ & SOD$\!\uparrow$ &
SR$\uparrow$ & TPR$\uparrow$ & ATR$\downarrow$ & SOD$\!\uparrow$ &
SR$\uparrow$ & TPR$\uparrow$ & ATR$\downarrow$ & SOD$\!\uparrow$ &
SR$\uparrow$ & TPR$\uparrow$ & ATR$\downarrow$ & SOD$\!\uparrow$ &
SR$\uparrow$ & TPR$\uparrow$ & ATR$\downarrow$ &SOD$\!\uparrow$ \\

DP~\cite{chi2023diffusionpolicy} &
\textbf{1.00} & 0.46 & 2.23 & 1.08 &
\textbf{0.97} & 0.18 & 5.52 & 1.05 &
0.83 & 0.10 & 7.56 & 0.99 &
\textbf{1.00} & 0.19 & 5.50 & 0.98 &
0.68 & 0.03 & 17.44 & 0.97 \\

DP-Fast &
0.95 & 1.02 & 1.52 & 1.59 &
0.87 & 0.37 & 2.34 & 2.48 &
0.55 & \textbf{0.15} & \textbf{3.42} & \textbf{2.20} &
0.98 & 0.44 & 2.37 & 2.28 &
0.56 & 0.05 & 9.67 & 1.74 \\

AWE~\cite{shi2023waypoint} &
\textbf{1.00} & 0.44 & 2.35 & 1.02 &
0.96 & 0.17 & 5.80 & 1.00 &
0.83 & 0.10 & 8.13 & 0.93 &
0.98 & 0.11 & 9.01 & 0.60 &
\textbf{0.75} & 0.02 & 28.79 & 0.59 \\

Agg.\ Act. &
0.91 & 0.52 & 1.78 & 1.37 &
0.82 & 0.16 & 4.77 & 1.22 &
0.29 & 0.03 & 4.81 & 1.57 &
0.82 & 0.17 & 6.18 & 0.87 &
0.59 & 0.03 & 15.86 & 1.06 \\

BID-Fast~\cite{liu2024bidirectional} &
0.86 & 0.91 & 0.97 & 2.50 &
0.79 & 0.34 & 2.39 & 2.34 &
0.49 & 0.12 & 3.45 & 2.18 &
0.99 & 0.47 & 2.61 & 2.07 &
0.62 & 0.06 & 8.71 & 1.94 \\

\textbf{SAIL(Ours)} &
\textbf{1.00} & \textbf{1.68} & \textbf{0.61} & \textbf{3.98} &
0.92 & \textbf{0.51} & \textbf{1.81} & \textbf{3.20} &
\textbf{0.86} & 0.13 & 6.41 & 1.18 &
0.98 & \textbf{0.66} & \textbf{1.56} & \textbf{3.47} &
0.72 & \textbf{0.08} & \textbf{8.09} & \textbf{2.09} \\

\end{tblr}
\end{table}

\subsection{Real-World Evaluation: SAIL Achieves High Throughput on Hardware}

\begin{table}[ht]
\centering
\caption{Real-World Evaluation}
\label{tab:real-eval_combined}
\scriptsize

\begin{tblr}{
  width       = \linewidth,
  colsep      = 2pt,
  colspec = {Q[l] | XXXX | XXXX | XXXX | XXXX },
  row{1}      = {font=\bfseries},
  hline{1,3,5}= {-}{},
  cell{1}{2}  = {c=4}{c},
  cell{1}{6}  = {c=4}{c},
  cell{1}{10} = {c=4}{c},
  cell{1}{14} = {c=4}{c},
}
 & Stacking Cups & & & &
   Wiping Board  & & & &
   Baking        & & & &
   Folding Cloth & & & \\
Method &
SR$\uparrow$ & TPR$\uparrow$ & ATR$\downarrow$ & SOD$\uparrow$ &
SR$\uparrow$ & TPR$\uparrow$ & ATR$\downarrow$ & SOD$\uparrow$ &
SR$\uparrow$ & TPR$\uparrow$ & ATR$\downarrow$ & SOD$\uparrow$ &
SR$\uparrow$ & TPR$\uparrow$ & ATR$\downarrow$ & SOD$\uparrow$ \\

DP-Fast &
0.10 & -2.28 & \textbf{14.00} & \textbf{1.85} &
\textbf{0.90} & \textbf{3.48} & 14.54 & 2.34 &
0.90 & 3.06 & 16.15 & 2.26 &
0.10 & -2.28 & 14.60 & 2.08 \\

\textbf{SAIL} &
\textbf{0.40} & \textbf{-0.12} & 14.71 & 1.76 &
0.70 & 3.18 & \textbf{10.44} & \textbf{3.26} &
\textbf{1.00} & \textbf{4.20} & \textbf{14.39} & \textbf{2.54} &
\textbf{0.30} & \textbf{-0.78} & \textbf{13.68} & \textbf{2.22} \\
\end{tblr}

\vspace{-1.5px}

\begin{tblr}{
  width       = \linewidth,
  colsep      = 2pt,
  colspec = {Q[l] | XXXX | XXXX | XXXX },
  row{1}      = {font=\bfseries},
  hline{1,3,5}= {-}{},
  cell{1}{2}  = {c=4}{c},
  cell{1}{6}  = {c=4}{c},
  cell{1}{10} = {c=4}{c},
}
 & Plate Fruits & & & &
   Pack Chicken & & & &
   Bimanual Serve & & & \\

Method &
SR$\uparrow$ & TPR$\uparrow$ & ATR$\downarrow$ & SOD$\uparrow$ &
SR$\uparrow$ & TPR$\uparrow$ & ATR$\downarrow$ & SOD$\uparrow$ &
SR$\uparrow$ & TPR$\uparrow$ & ATR$\downarrow$ & SOD$\uparrow$ \\

DP-Fast &
0.60 & 2.22 & 13.74 & 1.66 &
0.40 & 0.51 & 17.33 & 1.25 &
0.40 & 1.00 & 12.01 & 1.43 \\

\textbf{SAIL} &
\textbf{0.80} &\textbf{ 5.46} &  \textbf{8.53} & \textbf{2.67} &
\textbf{0.90} & \textbf{5.22} &  \textbf{9.40} &\textbf{ 2.30} &
\textbf{0.70} & \textbf{5.40} &  \textbf{7.19} & \textbf{2.39} \\
\end{tblr}
\vspace{-10pt}
\end{table}

\begin{figure*}[t]
    \centering
    \includegraphics[width=1.0\textwidth]{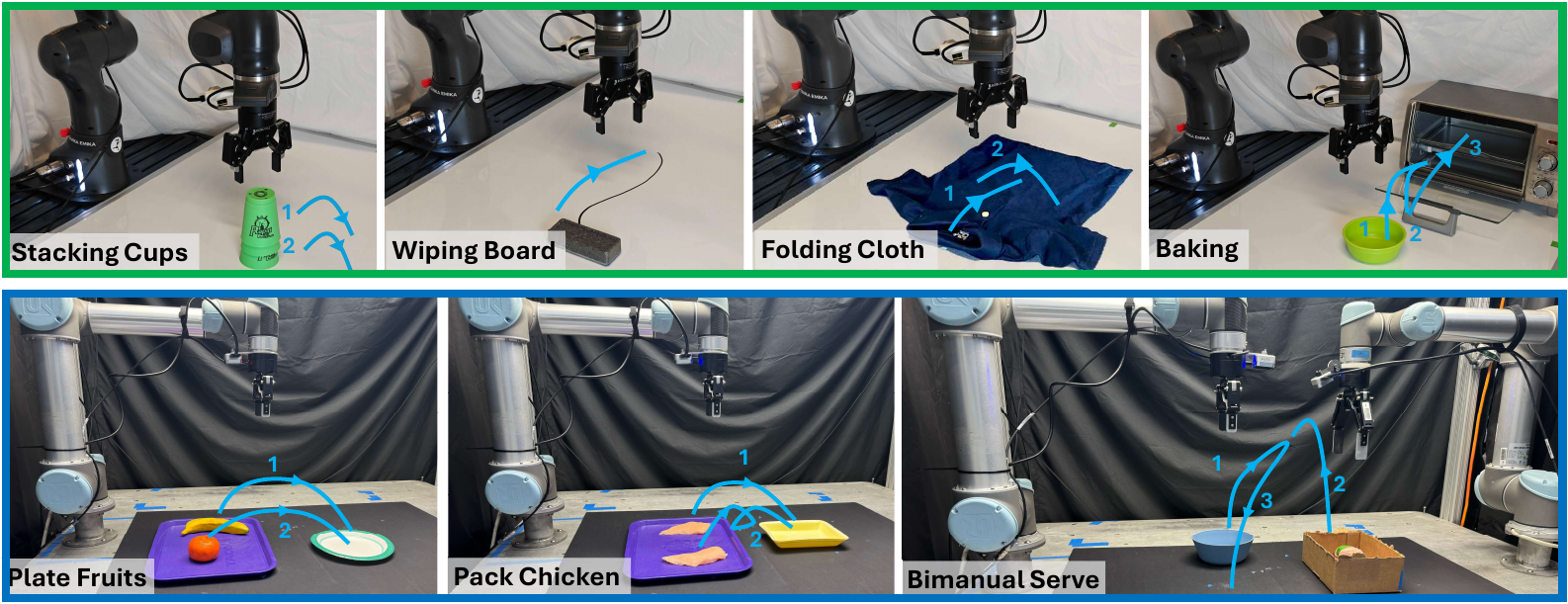} 
    \caption{Real-world task setup with two robot platforms (Franka and UR5).}
    \label{fig:real-tasks}
        \vspace{-10pt}
\end{figure*}

\textbf{Setup.}
We evaluate our method on two different robot platforms (Franka Emika Panda and Bimanual UR5) across 7 challenging manipulation tasks (shown in \Cref{fig:real-tasks}), including long task horizon, high-precision steps, and bimanual coordination (detail in \Cref{appendix:real-robot:task-description}).

Both robot platforms run high-fidelity tracking controllers (detail in \cref{appendix:sec:real_robot_setup}).
We compare \textbf{SAIL} ($5\times$ speedup) with \textbf{DP-Fast} ($5\times$ speedup). Each method is evaluated for 10 rollouts per task.

\begin{figure}[t]
    \centering
    \label{fig:failure_mode}
    \includegraphics[width=1.0\textwidth]{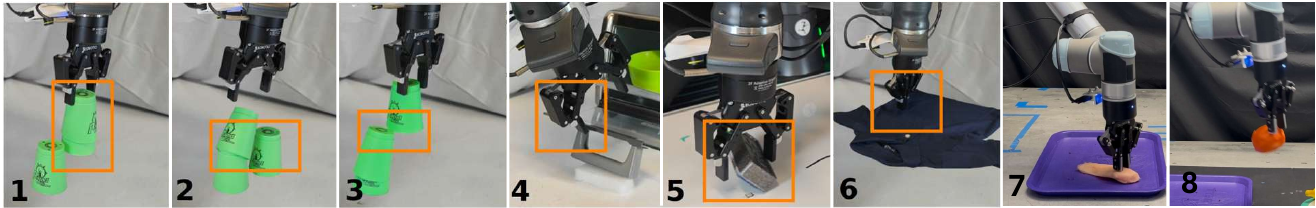} 
    \caption{\textbf{Commonly-seen failure modes from real-world evaluation.}
    Speeding up policy execution poses challenges unseen in normal speed execution, which include imprecise grasping (grasping two cups in 1, missing eraser in 5, and missing chicken in 7), low-fidelity tracking (colliding with other cups in 2 and 3, missing handle in 4, missing collar in 6), jerky motion (fruit drops in 8). SAIL effectively reduces such failures under speeding-up execution, leading to higher task throughput.}
    \vspace{-10pt}
\end{figure}

\textbf{Results: SAIL overcomes DP-Fast failure modes (\hypothesis{H3}).} Across differences in control systems, robot dynamics, and tasks, SAIL generally improves task throughput.
As shown in \Cref{tab:real-eval_combined}, throughput-with-regret (TPR) and SOD both improved on 6/7 challenging tasks, demonstrating SAIL's consistent speed advantage over the baseline sped-up diffusion policy.
Qualitatively (\Cref{fig:failure_mode}), SAIL overcomes common DP failure modes during high-speed execution.
DP often pauses due to \textbf{action depletion} when inference lags; SAIL maintains constant motion via smooth action scheduling. High speeds exacerbate \textbf{imprecise grasping} for DP, whereas SAIL's adaptive speed modulation slows critical phases, enhancing success rates in mid-to-high precision tasks such as plating fruits, packing chicken, and cup stacking. \textbf{Low-fidelity motion tracking} frequently causes DP failures, particularly for actions requiring precise localization like cloth folding and baking; SAIL substantially reduces tracking errors compared to the original teleoperation controller ($\controller\lbl{teleop}$), thus improving performance in precision-critical stages. Furthermore, DP's \textbf{inconsistent action predictions} result in jerky motions, posing considerable challenges at increased speeds. This is particularly evident in the bimanual serving task, where SAIL achieves smoother trajectories, yielding $5.4\times$ higher throughput and approximately $1.8\times$ greater success rates. Finally, we note that SAIL performs slightly worse than DP in the wiping task. We hypothesize that this is due to the high-gain tracking controller's inability to adjust to the new robot-object dynamics (sustained contact during wiping) at a higher speed. We provide additional discussion in the Limitation section.

\section{Conclusion}
\label{sec:conclusion}
We formalized and identified challenges in the novel problem of faster-than-demonstration execution of visuomotor policies.
Our framework, SAIL, tackles the full-stack problem by combining Error-Adaptive Guidance, controller-invariant targets, adaptive speed modulation, and latency-aware scheduling. Experiments show SAIL achieves up to 4$\times$ speedup in simulation and 3.2$\times$ speedup in real-world while maintaining high success rates across diverse tasks.

\section{Limitations}
Through formalizing and proposing a solution for the novel problem of faster-than-demonstration execution, we have identified several fundamental challenges that open up new research directions for the robotics community.
First, SAIL focuses on addressing the observation-action drift as a result of controller dynamics shift and does not explicitly tackle the dynamics shift of robot-object interaction.
This manifests most clearly in manipulation tasks where object-robot dynamics become significantly more complex at higher speeds---for instance, we observed that in the simulated Can task, increased execution speed can cause the robot to inadvertently throw the can out of the workspace due to increased momentum. Relatedly, the finite-data nature of offline imitation learning makes it vulnerable to distributional shift that cannot be addressed solely by improving the policy learning algorithms. 
Future research could address this by developing methods to incorporate explicit dynamics modeling into policies, either by leveraging known dynamics models or learning from simulation-based data during training.

As the field continues to advance learning-based manipulation in the wild, we believe a key focus should be to enable the robot learning system to co-optimize the low-level control and the anticipated dynamic effects of the predicted actions at different execution speeds. We hope this work can open new pathways and facilitate wider adoption of learned policies in real industrial applications.

\clearpage
\acknowledgments{The authors would like to acknowledge the State of Georgia and the Agricultural Technology Research Program at Georgia Tech for supporting the work described in this paper. We also acknowledge funding from the AI Manufacturing Pilot Facility project under Georgia Artificial Intelligence in Manufacturing (Georgia AIM) from the U.S. Department of Commerce Economic Development Administration, Award 04-79-07808, NSF CCF program, Award 2211815, and NSF Award 1937592.}


\bibliography{main}  

\clearpage
\appendix

\renewcommand\thefigure{\thesection.\arabic{figure}}
\setcounter{figure}{0}    

\renewcommand\thetable{\thesection.\arabic{table}}
\setcounter{table}{0}

\section{Table of Contents}
The Appendix contains the following content:
\begin{itemize}
    \item \textbf{Formulas and explanation of evaluation metrics} (Appendix~\ref{ssec:appendix:metrics}): this section explains the different metrics used in the experiments for this paper.
    \item  \textbf{Simulation experiment setup} (Appendix~\ref{appendix:sim_exp_details}): this section describes the task suite and simulator used for the simulation experiments, as well as specific changes we made to enable speedup. 
    \item \textbf{Real world setup and tasks} (Appendix~\ref{appendix:sec:real_robot_setup}): we detail our hardware and data collection setup and provide descriptions of the different tasks we used to evaluate SAIL.
    \item \textbf{Derivation of Action Interval} (Appendix~\ref{appendix:derivation_lower_bound}): we derive a theoretical bound for the speedup factor $\speedupfactor_t$ that still maintains continuous execution without pauses due to inference.
    \item \textbf{Modulating speed} (Appendix~\ref{subsec:appendix:adaptive_speed}): this section describes two different methods of detecting critical actions that allow SAIL to modulate the speedup factor $\speedupfactor$  during various stages of a task.
    \item \textbf{Experiments for justifying components of SAIL} (Appendix~\ref{appendix:exp_hypothesis1-3}): in this section we present mini experiments that give insights into different components of SAIL and how they affect performance.
    \item \textbf{Error adaptive guidance experiments} (Appendix~\ref{appendix:eag_exp}): we present a deep dive into EAG and the importance of varying guidance based on tracking error.
    \item  \textbf{Aggregating actions algorithm} (Appendix~\ref{ssec:appendix:baseline}): this section includes the implementation of the \textbf{Aggregated Actions} baseline in~\Cref{tab:simeval}.
    \item \textbf{Policy and controller parameters} (Appendix~\ref{appendix:hyperparameters}): this section lists out the different hyperparameters used for the policies and robot controllers.
    \item \textbf{Ablation and Action Chunking with Transformers results} (Appendix~\ref{appendix:subsec:ablation}) : we perform a detailed ablation of the different components of SAIL and present results using a different policy algorithm - ACT ~\cite{zhao2023learning}.
\end{itemize}

\section{Evaluation Metrics}
\label{ssec:appendix:metrics}
In this section, we describe the metrics that we used for evaluation in detail.
\begin{enumerate}
    \item \textbf{SR} (higher is better):
    The \textit{success rate} is the number of successfully-completed tasks divided by the total number of trials.

    \item \textbf{TPR} (higher is better):
    We propose \textit{throughput-with-regret} to reward faster successes while penalizing all failures equally:
    \begin{align}
        \TPR =
            \frac{1}{N}\sum_{i = 1}^{N}
                \left(
                    \left(\frac{1}{t_i}\cdot S_i\right) - 
                    \left(\frac{1}{t\lbl{max}}\cdot(1-S_i)\right),
                \right)
    \end{align}
    where $t_i$ is the \textit{duration} (i.e., total simulated clock time) of rollout $i$, $t\lbl{max}$ is the maximum amount of time allowed per trial, and $S_i$ is the \textit{success} of rollout $i$.
    That is, $S_i= 1$ if trial $i$ was successful and $0$ otherwise.
    We halt all trials if they have not succeeded before $t\lbl{max}$, and declare such trials as failures.

    \item \textbf{ATR} (lower is better):
    We record the \textit{average time for successful rollouts}, where we compute the average time in seconds only for the successful rollouts out of all trials. 

    \item \textbf{SOD} (higher is better):
    We report the \emph{speedup-over-demo}, which is the average length of a demo divided by ATR.
    SOD indicates how much one speeds up the execution of the imitation learning policy compared to the training demonstration.

    \item \textbf{CON} (lower is better):
    To evaluate our CFG and action conditioning approach, we quantify the \textit{consistency} between overlapping parts of consecutive action sequences, we measure the change in actions at the transition point, specifically, CON $= \hat\action_{\futureactionhorizon} - \action_{\horizon{\exec}+ \futureactionhorizon}$, following the notation of \Cref{subsec:method:cfg}.
    
    \item \textbf{SPARC} (higher is better):
    SPARC (linear and angular spectral arc length) is a smoothness metric that evaluates the arc length of the Fourier magnitude spectrum of a trajectory's speed profile~\cite{balasubramanian2015smoothness}. It is an extended version of Spectral Arc Length (SAL)~\cite{balasubramanian2012robust}.
    
    In SAL, the magnitude spectrum $\fouriermagnitudespectrum(\angularfreq)$ of the Fourier transform of a speed profile $\speedprofile_t$ is normalized by its DC value $\fouriermagnitudespectrum(0)$:
    
    \begin{equation}
        \hat{\fouriermagnitudespectrum}(\angularfreq) = \frac{\fouriermagnitudespectrum(\angularfreq)}{\fouriermagnitudespectrum(0)}
    \end{equation}
    
    SAL integrates the arc length of $\hat{\fouriermagnitudespectrum}(\angularfreq)$ over frequencies from $0$ up to a cutoff frequency $\freqcutoff$:
    
    \begin{equation}
        \text{SAL} \triangleq -\int_{0}^{\freqcutoff} \left[ \left(\frac{1}{\freqcutoff}\right)^2 + \left(\frac{d \hat{\fouriermagnitudespectrum}(\angularfreq)}{d\angularfreq}\right)^2 \right]^{\frac{1}{2}} d\angularfreq
    \end{equation}
    
    where the first term in the square root is used for frequency normalization, normalizing the arc length with respect to $\freqcutoff$. 
    SPARC refines SAL by adaptively selecting $\freqcutoff$  based on a chosen amplitude threshold $\overline{V}$ and an upper limit $\freqcutoff^{\max}$ as follows:
    \begin{equation}
        \freqcutoff \triangleq \min \Bigl\{\freqcutoff^{\max},\min \bigl\{\angularfreq, \hat{\fouriermagnitudespectrum}(r) < \overline{\fouriermagnitudespectrum} \;\; \forall r>\angularfreq \bigr\} \Bigr\}.
    \end{equation}
    
    We compute SPARC for a given speed trajectory in the following manner. First, we pad the trajectory with zeros ($K=4$) to increase the frequency resolution to accurately estimate the length of the arc. Next, we normalize the magnitude spectrum and apply an upper limit $\freqcutoff^{\max}=20$ and an amplitude threshold $\overline{\fouriermagnitudespectrum}=0.05$. We then compute the arc length of the normalized spectrum by summing the Euclidean distance between successive frequency-domain points. Finally, we multiply this sum by $-1$ to obtain larger values for smoother trajectories.
    
        \item \textbf{LDLJ} (lower is better):
        Log Dimensionless Jerk (LDLJ) is a smoothness metric that evaluates how quickly and drastically the motion accelerates or decelerates based on the third derivative of position, jerk.
        
        LDLJ can be written as~\cite{balasubramanian2015smoothness}:
        \begin{align}
            \text{LDLJ} \triangleq -\ln\left|
                    -\,\frac{\bigl(t_{2} - t_{1}\bigr)^{5}}{(v\lbl{peak})^2} 
                    \int_{t_{1}}^{t_{2}} \left|\frac{d^{2}v_t}{dt^{2}}\right|^{2}\,dt
                \right|
        \end{align}
        where $t_1$ and $t_2$ are the start and end times of the movement, $v_t$ is the speed at time $t$, and $v\lbl{peak} \triangleq \max_{t \in [t_1,t_2]} v_t$.
        
        In our work, we calculate LDLJ by setting $v\lbl{peak}$  as the peak speed within the trajectory. The speed \(v(t)\) is computed as the difference between the positions of successive points. After estimating the speed, we apply finite differences to approximate its second derivative with respect to time. The squared second derivative is then integrated over the movement duration, scaled by \(\frac{(t_{2} - t_{1})^5}{v_{\mathrm{peak}}^2}\), and the negative natural logarithm is applied to obtain the final LDLJ value.
        \item \textbf{WED} (lower is better):
        We calculate the \textit{Weighted Euclidean Distance}, a metric that is used in~\cite{liu2024bidirectional} to quantify the consistency between overlapping action segments. 
    \end{enumerate}

\section{Simulation Experiment Details}
\label{appendix:sim_exp_details}

\subsection{\textbf{Robot Control and Dynamics Considerations}}
We control robots in simulation using an OSC controller that takes absolute pose commands, except for the \textbf{aggregated actions} baseline, which uses delta pose commands. 
Additionally, to ensure that robot torque limits are not a bottleneck to speed up, we removed the joint torque limits of the Franka Emika Panda robot in Robosuite. For some of our baselines, removing torque limits resulted in worse performance. In these cases, we report their performance with torque limits.
In \Cref{tab:sim_controller_params}, we list the optimal controller and the upper bound of $c$ in adaptive speed modulation that SAIL uses in the simulated tasks.

\subsection{\textbf{Simulator and Data}}
We use Robosuite~\cite{robosuite2020} to simulate robots and their environments. Robosuite is built on Mujoco~\cite{todorov2012mujoco} and by default simulates two milliseconds (0.002 s) of real-world time every time the simulator is stepped forward.
In our problem setting, we consider the action interval $\dt$ as the number of simulation steps allowed for a robot controller to execute a given action.
Speeding up policy execution is thus to reduce the total number of simulation step taken to finish a task. The teleoperation data is collected at $\delta=0.05$ s ($20$ Hz).

For our simulation benchmark, we use three tasks from the Robomimic~\cite{mandlekarmatters} suite of tasks and two tasks from MimicGen~\cite{mandlekar2023mimicgen}. For the Robomimic tasks,  we train a separate policy for each task on 200 human demonstrations. For MimicGen tasks, we use 500 machine-generated demonstrations for each task.

The diffusion policy is trained with prediction horizon $\horizon=32$. In a receding-horizon manner, we execute 8 of the predicted actions before the next inference, and 4 more actions while inference is running to simulate sensing-inference delay.

\subsection{\textbf{Compute}}
We run all sim experiments on a compute cluster.
Each experiment uses a single A40 GPU, 8 CPU cores and 64GB of RAM.


\section{Real-World Evaluation Setup}
\label{appendix:sec:real_robot_setup}
In this section, we explain the real robot setup and data collection pipeline used in the paper.

\subsection{\textbf{Franka Robot}}
\label{appendix:subsec:franka}
 We have a four-level control hierarchy for the Franka robot. 
 In the first level, action chunks from policy inference are retrieved in a variable frequency and velocity approximation is performed, as introduced in \Cref{sec:problem}.
 
 Second, actions in each chunk are interpolated and scheduled by a computer (Intel NUC)  controlling the robot at 100Hz.

 Third, we use our OSC controller for the Franka, which is based on the Deoxys controller introduced in~\cite{zhu2022viola}. Given the desired 6D pose, and velocity, we calculate the pose error $\error_p$ and velocity error $\error_v$.
The computed torque $\torque$ sent to robot joints is
\begin{equation}
    \torque = \jacobian\trans\massmatrix(\gain_p \error_p + \gain_v \error_v),
\end{equation}
where $\jacobian$ is the Jacobian of the robot, $\massmatrix$ is the mass matrix represented in end-effector space.
The velocity target is computed by fitting and differentiated a spline over the predicted action trajectory.

 Fourth and finally, we leverage the torque control API from libfranka and calculate torque commands, which are sent to the on-board Franka controller at 500Hz.

\begin{figure}[t!] 
    \centering
    \includegraphics[width=0.8\columnwidth]{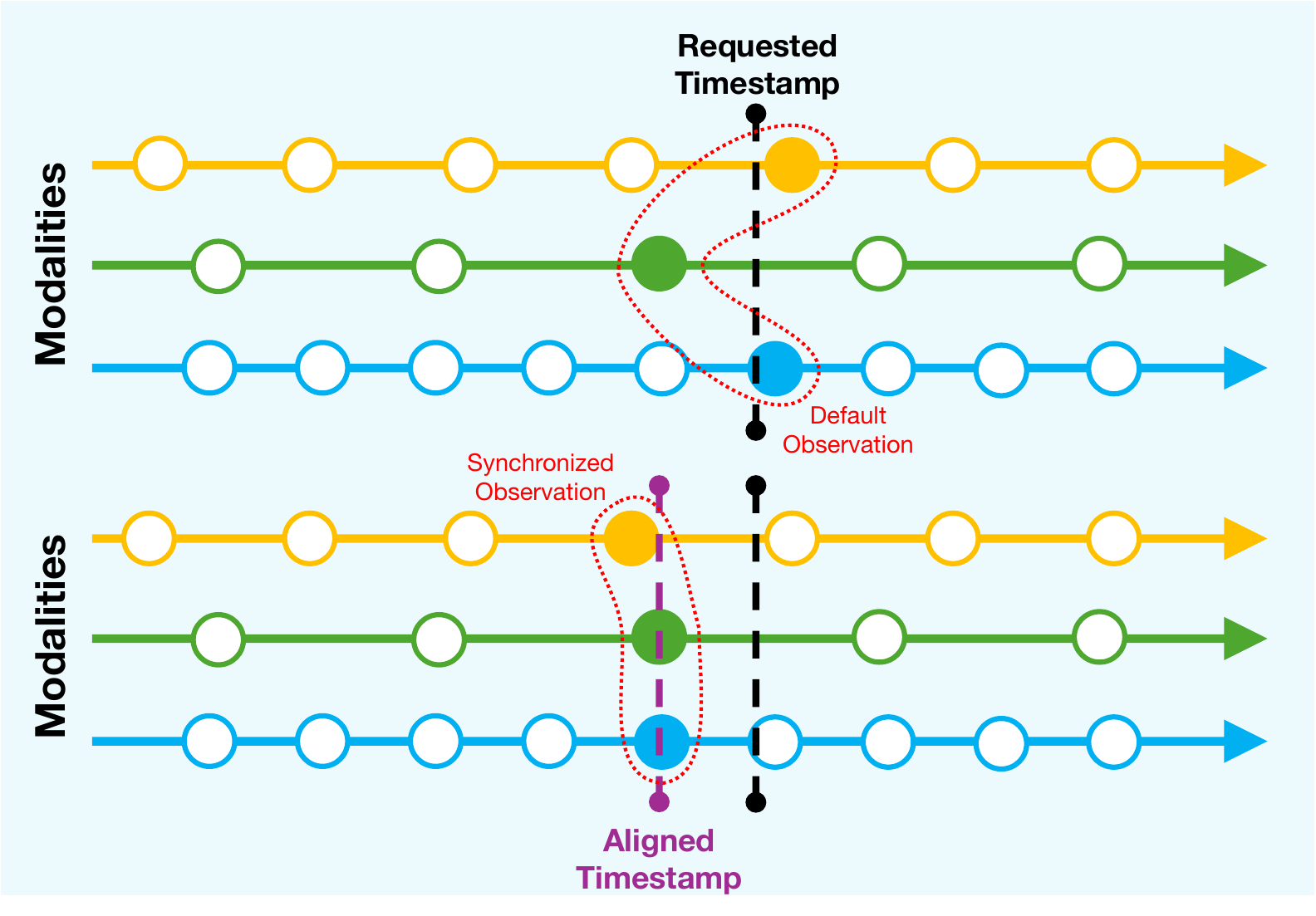}
    \caption{\textbf{Timestamps alignment}. For each modality, we identify the nearest observation (filled circles) to construct the default observation (red dotted outline) at the requested timestamp (black dashed line). Among these nearest timestamps, we determine the farthest one (purple dashed line). We then align the other modalities (filled circles) to this timestamp to obtain the synchronized observation (red dotted outline).}
    \label{fig:timestamps_alignment} 
\end{figure}

Our teleoperation system uses a Meta Quest VR headset to control the \textit{commanded pose} of the robot end effector.
To record an initial pose during the demonstration collection, users press the VR controller grip button. While the button is held, the change in the VR controller pose relative to the initial pose is transformed into the robot coordinate frame and used to adjust the \textit{commanded pose}.
Releasing the grip button pauses the teleoperation, allowing users to reposition their hand comfortably before resuming the task.
Demonstrations are recorded at 20Hz.
We collect the robot's \textit{reached poses} and \textit{commanded poses} at 20Hz.
Images are attained from the Kinect camera which is collected at 30Hz, while Zed camera is collected at 60Hz. To handle this mismatch, we cache the most recent 100 messages per modality and perform an observation alignment process. Specifically, upon receiving a request for observation at time $t$, we find the closest timestamps in each cache, then among these we identify the one farthest from $t$, and realign all modalities to that time to ensure consistency (see \Cref{fig:timestamps_alignment}).

We collect 50 demonstrations for each task.

\subsection{\textbf{UR5 Robot}}
We designed a modular control and data collection framework for UR5 robots equipped with Cartesian controllers~\cite{cartesianmotioncontroller} based on ROS 2\footnote{\url{https://docs.ros.org/en/jazzy/index.html}}.

Similar to the Franka setup, demonstrations are collected through a teleoperation loop running at 20Hz, while our observation modalities operate at different frequencies: the on-wrist Intel RealSense D405 and static D435 scene cameras run at 90Hz, the robot controllers operate at 125Hz, and the VR device at 72Hz. 

By scheduling actions our system can execute the learned policies at configurable rates. By default, we match the 20Hz demonstration rate, but the frequency can be dynamically adjusted based on model predictions to speed up or slow down motions. Our UR5 setup can be used to perform bimanual tasks with a single operator/policy controlling both arms. Each arm's speed can be individually controlled as needed to complete the task. For the tasks tested on the UR5 robots, we varied the number of demos according to the task length and difficulty.

\subsection{\textbf{Task Descriptions}}
\label{appendix:real-robot:task-description}
\begin{itemize}
    \item \textbf{Stacking Cups}.
    This task mimics speed stacking, wherein humans attempt to stack cups in predetermined sequences as quickly as possible.
    We collect human teleoperation demos to stack 3 cups into a pyramid shape for this task.
    The repetitive grasping, placing, and movement pose challenges in balancing speeding up policy and manipulation accuracy.
    
    \item \textbf{Baking}.
    The baking task represents use cases in a commercial kitchen where efficiency are important.
    The goal for this task is to pick up a bowl from a table and place it precisely on an oven rack.
    Then the robot must close the oven door, which requires precise contact-rich motion to accomplish.
    
    \item \textbf{Folding Cloth}.
    This task requires the robot to fold a t-shirt.
    The robot must pick the collar and fold the whole shirt in half, then pick the right sleeve to do another fold.
    Success requires finishing both folds.
    
    \item \textbf{Wiping Board}.
    To showcase SAIL's robustness we include a contact-rich manipulation scenario: the wiping-board task. The robot must pick up an eraser and wipe a line on a whiteboard, all while maintaining forceful contact.
    Since the demonstration data does not include any force and contact information, accelerating execution of such a task with imitation learning is challenging.

    \item \textbf{Plate Fruits}.
    This task consists of picking two fruits from a random position on a tray and placing them on a plate with a specific pattern. This task is challenging because the picking order is fixed, but the positions can be swapped. Policies for this task were trained with 100 demos.

    \item \textbf{Pack Chicken}.
    This real world packing task is challenging because of the variability in the shape of the deformable (rubber) chicken breasts and the limited size of the container requiring precise placements. The robot is required to rotate the second chicken breast to fill the space in the container. This complex motion makes speeding up challenging. Policies for this task were trained with 100 demos.

    \item \textbf{Bimanual Serve}.
    This task involves two robots operating together. While the first is picking a peach, the second is picking a bowl. Then, they converge to a common point where the first robot places the peach in the bowl. Then the bowl with the peach inside is served. This task is very difficult to accelerate because it requires alignment and synchronization between the two arms. Policies for this task were trained with 75 demos.
\end{itemize}

\section{Derivation of Lower Bound \texorpdfstring{$\dtbound$}{dtbound} for Action Interval}
\label{appendix:derivation_lower_bound}
We seek to derive a lower bound for the action interval $\dt$ (i.e., highest speed up) that still allows a continuous control loop. Recall that $\horizon\plan$ is the prediction horizon of the policy, and let $\dtbound$ be the (constant) lower bound on $\dt_t$ that we aim to find.
To find it, we consider the three critical parameters:
(a) Sensing-inference delay $\dt\delay = t^\action - t^\observation$,
(b) the shortest-possible action execution time $\horizon\plan\cdot\dtbound$, and
(c) the shortest-possible length of the action chunk used for consistency-preserving conditioning $\horizon\cond$ (\Cref{subsec:method:cfg}). 
To ensure continuous execution, we require the available action sequence to be longer than the sensing-inference delay plus the conditioning horizon, which gives the lower bound $\dtbound$ as
\begin{align}
\begin{split}
    &\horizon\plan \cdot \dtbound > \dt\delay + H\cond\dtbound  \\
    \implies\quad
    &\dtbound > \frac{\dt\delay}{\horizon\plan - \horizon\cond}.
\end{split}
\end{align}

\begin{figure}
  \centering
  \includegraphics[width=0.75\textwidth]{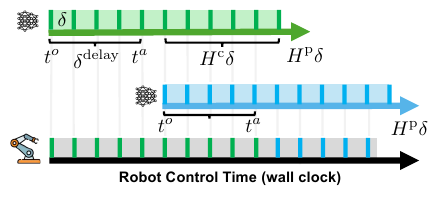}
  \caption{\textbf{Handling latencies in the control loop.}
    We illustrate the control loop timeline of SAIL and how it handles system latency. The green timeline (top) shows the first action sequence generated at $t^\observation$. The sequence spans $H\dt$ with the last $H\cond$ steps conditioning the next prediction. The blue timeline (middle) shows the next action prediction starts while the system continues to execute the first prediction. The bottom timeline shows the actual robot execution timeline. The system smoothly transitions from the first sequence (green) to the next (blue) without pausing.} 
  \label{fig:action_scheduling}
\end{figure}

We illustrate this relationship in \Cref{fig:action_scheduling}. 
Note that we can reduce $\dtbound$ (i.e., allow higher speedup) by extending the prediction horizon $\horizon\plan$, but this would require accurate action prediction over a longer horizon, which is inherently challenging~\cite{janner2021offline}.
Hence, in practice, the prediction horizon $\horizon\plan$ and the sensing-inference delay $\dt\delay$ jointly determines the minimum bound on $\dt_t$ and in turn the speedup factor $\speedupfactor_t$:
\begin{align}
    \speedupfactor_t \in \left(
            \dtbound / \dtoriginal,\, 1
        \right].
\end{align}
Moreover, note that $\speedupfactor_t$ from \Cref{subsec:adaptive_speed_execution} does not affect this computation, since the lower bound provides a worst-case guarantee---even though $\speedupfactor_t$ may increase to slow down execution in certain phases, we cannot rely on this a priori when computing $\dtbound$ for continuous execution.

\section{Adaptive Speed Modulation}
\label{subsec:appendix:adaptive_speed}

Specifically, we aim to identify critical actions in demonstrations, train the policy to predict both the actions and their critical action labels, and dynamically the speed up factor $c$ accordingly during execution.

\begin{figure}[t!] 
    \centering
    \includegraphics[width=1.0\columnwidth]{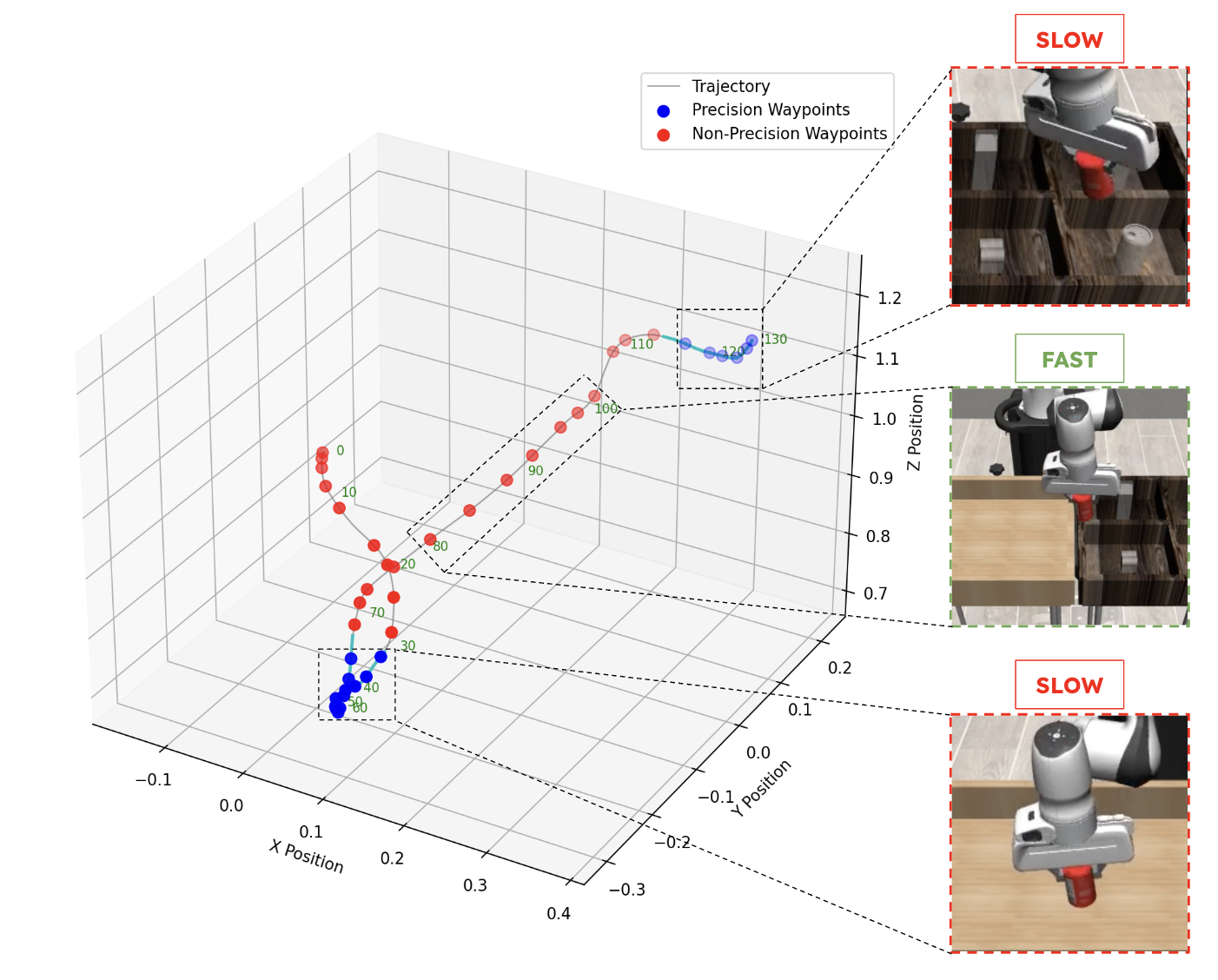}
    \caption{\textbf{Policy rollout with adaptive speed modulation} .
    Waypoints (red) are generated by~\cite{shi2023waypoint} given the trajectory as input along with an error threshold. Areas of complex motion (blue), are marked by performing a spatial clustering of the extracted waypoints. Any waypoint outside the cluster threshold we label as noise. Frame numbers are labeled every 10 steps in green -- one can observe that the clustering has been performed properly by the increased concentration of frame numbers in clustered areas (the end effector spends more time in these regions).
    }
    \label{fig:awe} 
\end{figure}

We propose two techniques to identify critical actions from the demonstration data: measuring motion complexity and using gripper open/close actions.
Each technique returns a binary \textit{critical action flag} $\critactflag_t \in \{0,1\}$ ($\critactflag_t = 1$ means that $\action_t$ is critical), which we use to set the speedup factor $\speedupfactor_t$. Given the binary critical action flag $\critactflag_t$, we set the speedup factor as:
\begin{equation}\label{eq:controlspeeddef}
    \speedupfactor_t = \critactflag_t\cdot \speedupfactor\slow + (1-\critactflag_t)\cdot \speedupfactor\fast.
\end{equation}
where $\speedupfactor\slow \in (0,1]$ is a slower speedup factor used for critical actions and $\speedupfactor\fast \in (0,1]$ is used otherwise.
Note that $\speedupfactor\slow > \speedupfactor\fast$ since the \textit{reciprocal} of the speedup factor determines the speedup.
We set $\speedupfactor\slow$ and $\speedupfactor\fast$ to empirically-validated presets for each task. 
Also note that we lower-bound $\speedupfactor_t$ in \Cref{appendix:derivation_lower_bound}.
A policy rollout using this technique of adaptive speed modulation is shown in \Cref{fig:awe}.

\textbf{Identifying Critical Actions via Motion Complexity.}
Inspired by Automatic Waypoint Extraction (AWE)~\cite{shi2023waypoint}, we approximate the commanded robot end effector poses in a demonstration with a set of waypoints connected by linear segments.
With a set error budget, this algorithm produces more waypoints for more complex motion.
We then identify fast and slow regions by clustering the waypoints in 3-D space using DBSCAN~\cite{mester1996dbscan}, which is well-known to identify arbitrarily-sized clusters better than spherical or centroid-based clustering methods like k-means; it also implicitly filters out noisy data points that may not represent significant motion changes.
To correlate each time step $t$ with a waypoint, we linearly interpolate between the waypoints in time.
Finally, given a minimum cluster size, we label each time-interpolated waypoint with $\critactflag_t = 1$ if the waypoint at time $t$ is in a cluster and $\critactflag_t=0$ otherwise.

We describe the algorithm to classify critical actions via motion complexity in \Cref{alg:awe_alg}.

\begin{algorithm}
\caption{Identify Critical Actions via Motion Complexity}\label{alg:awe_alg}
\KwIn{Absolute action sequence $\mathcal{A}=\langle a_{1},a_{2},\dots,a_{N}\rangle$}

\SetKwFunction{AWE}{AWE}
\SetKwFunction{SpeedLabel}{SpeedLabel}
\SetKwProg{Fn}{Function}{\string:}{}

\medskip
\Fn{\AWE{$\mathcal{A},\,\tau$}}{%
  \Return waypoint set $\mathcal{W}=\{w_{1},\dots ,w_{M}\}$ with $w_k=(x_k,y_k,z_k)$, threshold $\tau$\;
}
\Fn{\SpeedLabel{$\mathcal{W, \varepsilon, \textit{minPts}}$}}{%
  Run DBSCAN on $\{(x_k,y_k,z_k)\}_{k=1}^{M}$ with parameters $(\varepsilon,\textit{minPts})$\;
  \For{$k\gets 1$ \KwTo $M$}{
      \lIf{\textnormal{$w_k$ is assigned to a cluster}}{$\mathbf{label}[k]\gets 1$}
      \lElse{$\mathbf{label}[k]\gets 0$}
  }
  \Return label vector $\mathbf{label}$\;
}
$\mathcal{W} \gets$ \AWE($\mathcal{A},\,\tau$)\;
$\mathbf{label} \gets$ \SpeedLabel($\mathcal{W},\,\varepsilon,\,\textit{minPts}$)\;
$\mathbf{s}\gets\mathbf{0}_{N}$\;
\ForEach{consecutive pair $(w_{k},w_{k+1})$}{
    $i\leftarrow\text{index}(w_{k})$\;
    $j\leftarrow\text{index}(w_{k+1})$\;
    \If{$\mathbf{label}[k]=1$ \textbf{and} $\mathbf{label}[k+1]=1$}{
        \For{$t\gets i$ \KwTo $j$}{ $\mathbf{s}[t]\gets 1$\; }
    }
}

\BlankLine
\For{$i\gets 1$ \KwTo $N$}{
    $\tilde{a}_{i}\leftarrow (a_{i},\, \mathbf{s}[i])$\;
}
$\tilde{\mathcal{A}}\leftarrow\langle \tilde{a}_{1},\dots,\tilde{a}_{N}\rangle$\;

\KwOut{Augmented sequence $\tilde{\mathcal{A}}=\langle (a_{1},s_{1}),\dots,(a_{N},s_{N})\rangle$}
\end{algorithm}

\textbf{Identifying Critical Actions via Gripper Events}.
We observe that motions involving critical actions often occur during interactions with objects and the environment, which are correlated with gripper state changes.
Thus, we use these \textit{gripper events} to identify the critical actions.
That is, we set $\critactflag_t = 1$ if the gripper is changing (opening or closing) at $t$ and $\critactflag_t=0$ otherwise.

\section{Experiments for Testing Hypothesis for Each Component}
\label{appendix:exp_hypothesis1-3}
\subsection{Testing \hypothesis{H1}: Speeding Up Policy Execution Requires a High-Gain Controller}\label{subsec:exp1}

\begin{figure}[htbp] 
    \centering
    \includegraphics[width=\columnwidth]{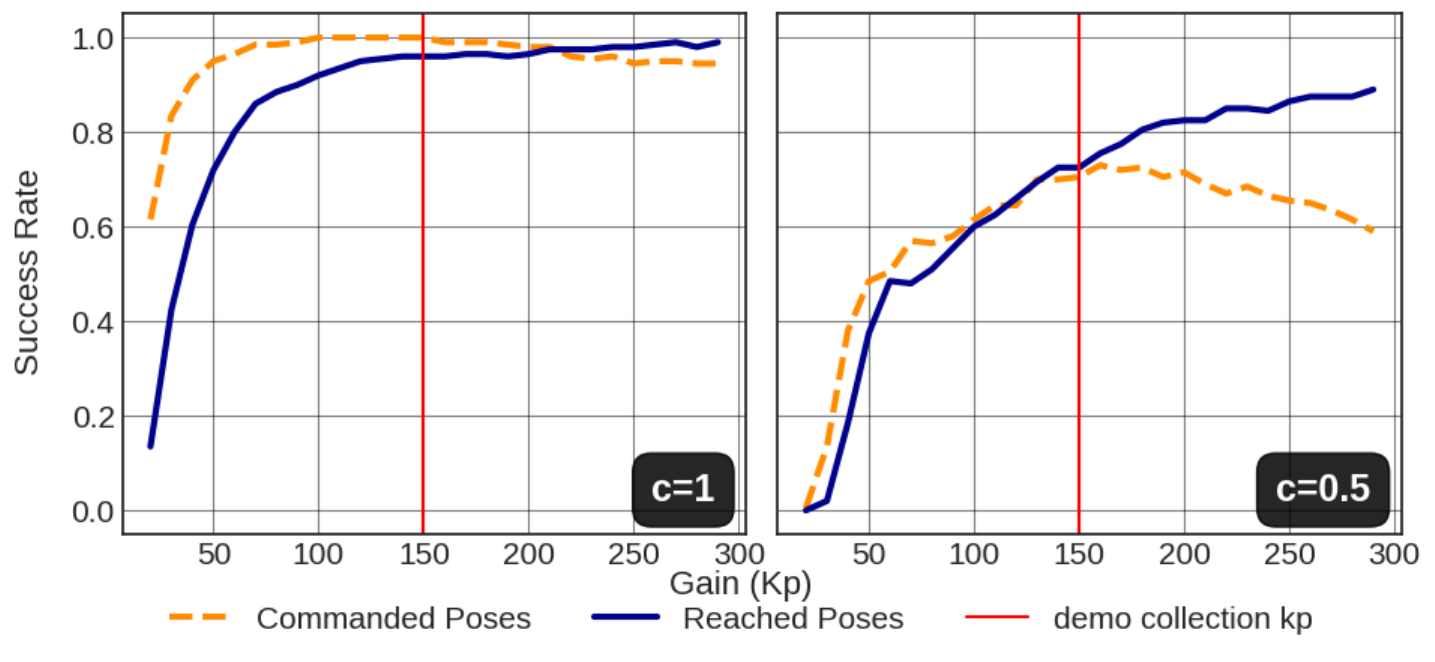}
    \caption{\textbf{Demo replay at different speeds and controller gains.} 
    We examine the effects of increasing controller gains and speed for replaying demos in simulation. Left: using commanded poses performs better when replaying at the original speed ($\speedupfactor = 1$) but using reached poses matches performance when using high gains.
    Right: A high-gain controller using reached poses performs better than one using commanded poses at a higher execution speed.
    }
    \label{fig:replay_over_kp} 
\end{figure}

To illustrate how controller tracking performance affects task execution during runtime, we replay the demonstrations of the Can task at different speeds and record the success rate for a given controller gain ($K_p$). We compare replaying the commanded poses ($\state\des$) and reached poses ($\state$) in demonstrations. As shown in \Cref{fig:replay_over_kp}, \textbf{High-gain control combined with reached pose tracking enables consistent behavior across execution speeds.}. Commanded poses lead to overshooting and task failures when attempting faster execution with higher gains. In contrast, using reached poses as reference trajectories enables consistently high success rates at increased speeds, provided the controller gain is sufficiently high to ensure accurate tracking.

\subsection{Testing \hypothesis{H2}: A High-Gain Controller Requires a Smooth Reference Trajectory}
\label{subsec:exp2}
In this experiment, we assess task success rate versus increasing noise scale.

This is important because faster execution can result in out-of-distribution observations that cause a policy to produce noisier reference trajectories.

\begin{figure}[htbp] 
    \centering
    \includegraphics[width=0.8\columnwidth]{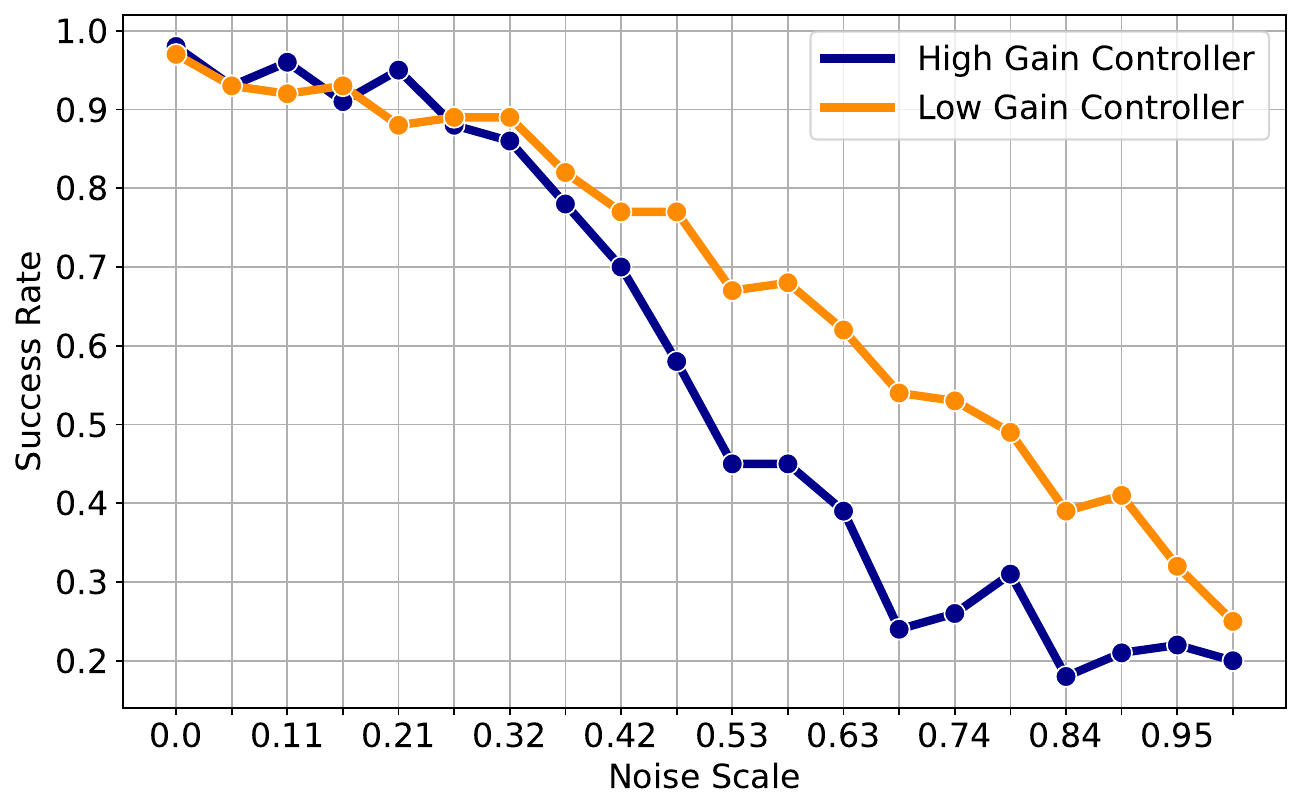}
    \caption{\textbf{Noise vs high-gain and low-gain controller.}
    We study the effects of increasingly noisy actions with a high-gain and low-gain controller. Success Rate is averaged over 100 rollouts per noise scale. At higher noise levels, the success rate when using a high-gain controller drops more significantly than with a low-gain controller. 
    }
    \label{fig:gain_vs_noise} 
\end{figure}

\textbf{High-gain controllers are more sensitive to noisy reference trajectories.}
As seen in \Cref{fig:gain_vs_noise}, as the noise scale increases, the high-gain controller's performance deteriorates more rapidly than the low-gain controller. This reveals an important coupling in our system: while high-gain control is necessary for accurate tracking at increased speeds, it also amplifies any inconsistencies in the predicted reference trajectories. This explains the need for trajectory smoothing in our proposed method.

\subsection{Testing \hypothesis{H3}: Action Conditioning Improves the Temporal Consistency of Across Predictions}
\label{subsec:exp3}
Since inconsistent action predictions are the main reason for unsmooth reference trajectories and failure in faster execution, we need to test whether action conditioning can improve temporal consistency and smoothness of generated actions.
To quantify this improvement, we conduct rollouts with and without action conditioning and evaluate smoothness and consistency using the SPARC, CON, and BID metrics (see \Cref{ssec:appendix:metrics} for details). We assess these metrics across five tasks at a speed-up factor of 0.2. As delineated in \Cref{tab:smoothness}, we found that the action conditioning results in smoother actions, supported by the higher SPARC metric and lower CON, BID metric compared to baseline DP.

\begin{table}[]
  \caption{\textbf{Evaluation of smoothness of actions}. We compare the smoothness of the generated actions using our method (c=0.2)}
  \centering
  \scriptsize          
  \renewcommand{\arraystretch}{1.0}  
  \setlength{\tabcolsep}{3pt}        
  \begin{tabular}{|ll|llll|}
    \hline
    \multicolumn{2}{|l|}{} & \multicolumn{1}{|l|}{SR $\uparrow$} & \multicolumn{1}{l|}{SPARC $\uparrow$} & \multicolumn{1}{l|}{CON $\downarrow$} & \multicolumn{1}{l|}{WED $\downarrow$} \\
    \hline
    
    \multicolumn{1}{|l|}{\multirow{3}{*}{Lift}} 
      & SAIL      & \textbf{0.97}  & \textbf{-2.80} & \textbf{0.091} & \textbf{0.348}  \\ \cline{2-2}
    \multicolumn{1}{|l|}{}   
      & BID~\cite{liu2024bidirectional}      & 0.53         & -2.85        & 0.116         & 0.395 \\ \cline{2-2}
    \multicolumn{1}{|l|}{}                               
      & Baseline  & 0.92         & -2.83        & 0.094         & 0.376  \\ \hline
      
    \multicolumn{1}{|l|}{\multirow{3}{*}{Can}}  
      & SAIL      & \textbf{0.76} & \textbf{-2.80} & 0.232         & 0.885 \\ \cline{2-2}
    \multicolumn{1}{|l|}{} 
      & BID~\cite{liu2024bidirectional}      & 0.62         & -2.93        & 0.325         & \textbf{0.525}  \\ \cline{2-2}
    \multicolumn{1}{|l|}{}                               
      & Baseline  & 0.73         & -2.83        & \textbf{0.230} & 0.930 \\ \hline
      
    \multicolumn{1}{|l|}{\multirow{3}{*}{Square}}        
      & SAIL      & \textbf{0.87} & \textbf{-2.74} & \textbf{0.107} & 0.916 \\ \cline{2-2}
    \multicolumn{1}{|l|}{} 
      & BID~\cite{liu2024bidirectional}      & 0.12         & -3.06        & 0.217         & \textbf{0.327}  \\ \cline{2-2}
    \multicolumn{1}{|l|}{}                              
      & Baseline  & 0.81         & -2.80        & 0.121         & 0.957  \\ \hline
      
    \multicolumn{1}{|l|}{\multirow{3}{*}{Stack}} 
      & SAIL      & 0.90         & \textbf{-2.50} & 0.168         & \textbf{0.634}  \\ \cline{2-2}
    \multicolumn{1}{|l|}{} 
      & BID~\cite{liu2024bidirectional}      & \textbf{0.92} & -2.56        & 0.641         & 0.794  \\ \cline{2-2}
    \multicolumn{1}{|l|}{}                               
      & Baseline  & 0.89         & -2.56        & \textbf{0.156} & 0.739  \\ \hline
      
    \multicolumn{1}{|l|}{\multirow{3}{*}{Mug}} 
      & SAIL      & \textbf{0.63} & -2.80        & \textbf{0.228} & 1.192  \\ \cline{2-2}
    \multicolumn{1}{|l|}{} 
      & BID~\cite{liu2024bidirectional}      & 0.40         & \textbf{-2.62} & 0.831         & \textbf{0.679}  \\ \cline{2-2}
    \multicolumn{1}{|l|}{}                               
      & Baseline  & 0.59         & -2.88        & 0.276         & 1.210  \\ \hline
      
  \end{tabular}
  \label{tab:smoothness}
\end{table}

\section{EAG Experiments}
\label{appendix:eag_exp}
We first describe Error Adaptive Guidance in \cref{alg:EAG}.

\begin{algorithm}
\caption{Error Adaptive Guidance for Diffusion Policy}\label{alg:EAG}
\KwIn{Diffusion Policy $\policy$, observations $\observation$, future actions $\conditioningactions$,
    current end-effector pose $(\mathbf{x}_{\mathrm{pos}},\,\mathbf{x}_{\mathrm{ori}})$, 
    desired pose $(\mathbf{a}_{\mathrm{pos}},\,\mathbf{a}_{\mathrm{ori}})$, error thresholds ($pos_{teb}$, $ori_{teb}$)}

  \SetKwFunction{ComputeTrackingError}{ComputeTrackingError}
  \SetKwProg{Fn}{Function}{:}{\KwRet}
  \SetKwComment{Comment}{$\triangleright$ }{}
  \Comment{Calculate tracking error}
  
      $\Delta\mathbf{p} = \mathbf{a}_{\mathrm{pos}} - \mathbf{x}_{\mathrm{pos}}$\;
      $e_{\mathrm{pos}} = \lVert \Delta\mathbf{p} \rVert$\;
      $R_{\triangle} = R(\mathbf{a}_{\mathrm{ori}})^{\top}\,R(\mathbf{x}_{\mathrm{ori}})$\;
      $e_{\mathrm{ori}} = \arccos\!\Bigl(\dfrac{\operatorname{tr}(R_{\triangle})-1}{2}\Bigr)$\;

  \Comment{Add action conditioning to observations}
    \lIf {$e_{\mathrm{pos}} > pos_{teb}$ \textbf{or} $e_{\mathrm{ori}} > ori_{teb}$} 
    {$\observation \gets \observation \cup \emptyset$}
    \lElse{$\observation \gets \observation \cup \conditioningactions$}
  \Comment{Generate next set of actions}
  $\mathcal{A}  \gets \policy(\observation)$ \;
\KwOut{Action sequence $\mathcal{A}=\langle a_{1},a_{2},\dots,a_{N}\rangle$}
\end{algorithm}

Next, we validate the key insights that motivated the design of consistency-preserving action prediction generation. 
Specifically, we test the following hypotheses:
\begin{itemize}
    \item \hypothesis{H-EAG1}: Consistency-guiding is beneficial when action condition is aligned with observation
    \item \hypothesis{H-EAG2}: Policy speed-up results in more misalignment between observation and action condition
    \item \hypothesis{H-EAG3}: Tracking error is highly correlated to observation-action misalignment
\end{itemize}

\subsection{Testing \hypothesis{H-EAG1}: Consistency-guiding is beneficial when action condition is aligned with observation}
\label{appendix:subsec:CFG_experiments}

We provide an additional study on the key factors influencing our Classifier-Free Guidance (CFG) approach for preserving action consistency.
Namely, we test the hypothesis \hypothesis{H-EAG1}: consistency guiding is effective when conditioned on future action condition in unconditional action distribution.

\begin{figure*}[htbp] 
\centering \includegraphics[width=\textwidth]{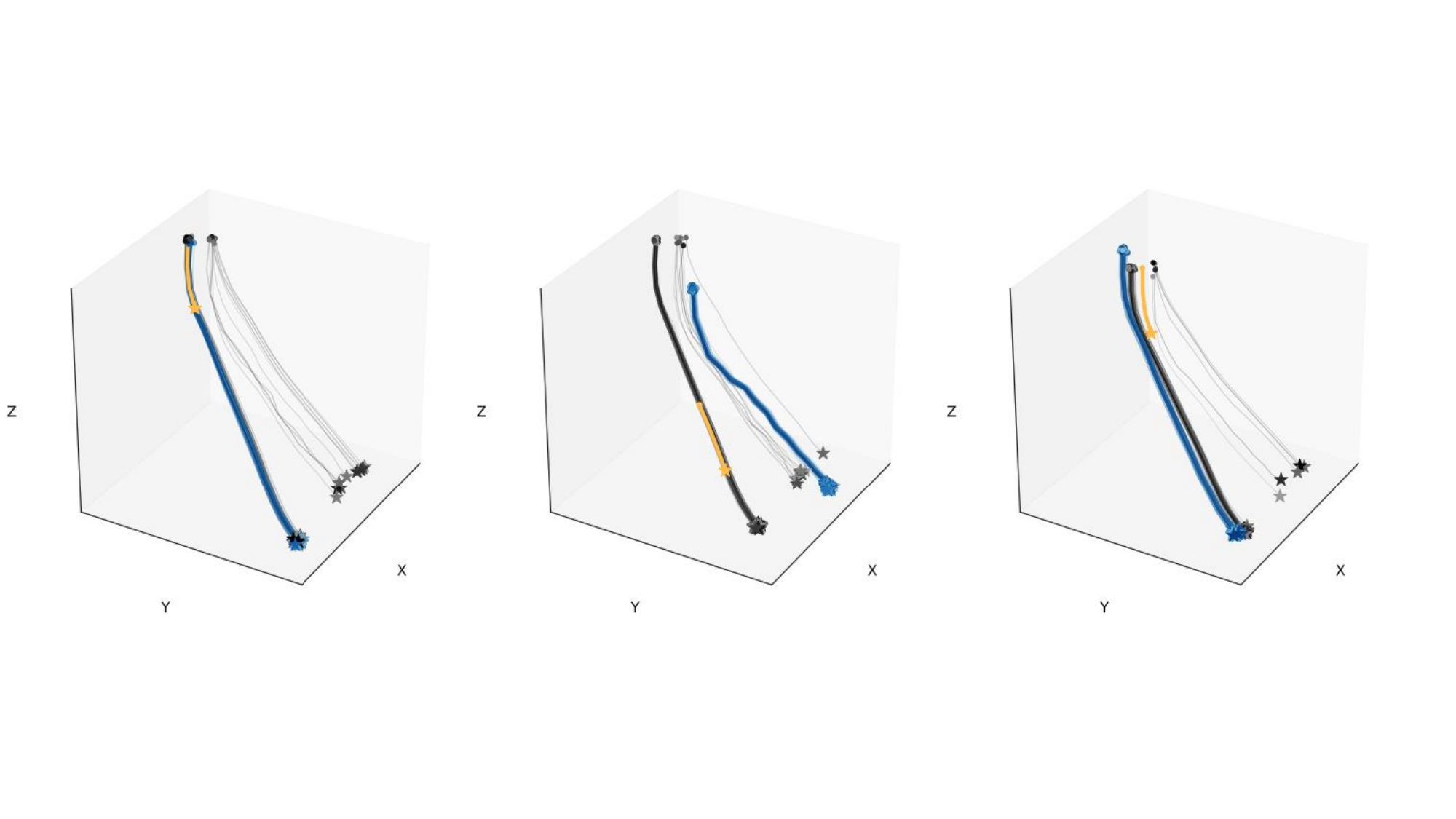} 
\vspace{-2.5cm} 
\caption{
    \textbf{Impact of Action Conditioning on Consistency Guidance.} 
    We evaluate how our algorithm adheres to conditioning on predicted actions.
    The action conditioning is shown in yellow; samples from the unconditional action distribution are shown in grey; and the conditional action distribution is shown in blue.
    The left panel shows future action conditions sampled from the unconditional distribution.
    The center panel shows temporally shifted future action conditions, while the right panel shows spatially perturbed future action conditions.
    When the action conditioning (yellow) lies within the unconditional action distribution (grey) in both space and time, the conditional action distribution (blue) is consistent with the action conditioning.
    However, when the action conditioning falls outside the unconditional action distribution, the model struggles to align the generated actions with the given condition.
    The middle subplot shows time-shifted action conditioning and the right subplot shows space-shifted action conditioning.
} 
\label{fig:fac_vs_uad} 
\end{figure*}

In this experiment, we investigate when consistency guidance is beneficial, specifically when the future action condition belongs to the unconditional action distribution. We evaluate how well the generated actions conform to the future action condition under different perturbations.

Our key argument is that consistency guidance is most effective when the future action condition exists within the unconditional action distribution. To validate this, we conduct the following experiment. Given a fixed observation, we sample 64 action sequences from the unconditional action distribution by setting the future action condition to a null token. Next, we select one of these samples as the future action condition and generate 64 action sequences using Classifier-Free Guidance (CFG) with a weight of 1. The resulting conditional action distribution is shown in the left column of \Cref{fig:fac_vs_uad}.

To analyze the effects of perturbations, we apply two modifications. First, we introduce a temporal shift by delaying the actions, as shown in the center column. Second, we introduce spatial perturbations by adding uniformly sampled noise (range: 0.02) to the selected actions, as shown in the right column.

Our results indicate that CFG performs best when the future action condition is within the unconditional action distribution. This suggests that such conditions are likely present in the training dataset, meaning the model has encountered similar action-observation pairs during training. Consequently, the model learns to correctly condition on these actions. However, when the future action condition deviates from the unconditional distribution—potentially due to tracking errors—the model may struggle to compute an appropriate conditional score.

\subsection{Testing \hypothesis{H-EAG2}: Policy speed-up results in more misalignment between observation and action condition}
\label{appendix:subsec:policy_speed_vs_fac}
We now examine how policy speed-up influences action conditioning.
Specifically, we hypothesize that increasing policy speed-up results in the action condition deviating further from the unconditional action distribution, potentially degrading the performance of CFG.

To validate this, we conduct the following experiment. We construct a batch of scenarios consisting of the simulation's internal state and corresponding observations from expert demonstrations.
For each scenario, we reset the simulation to the recorded internal state and use the diffusion policy to generate actions $\action_{0:\horizon}$. 
Following the notation in \Cref{subsec:method:cfg}, we execute $\action_{0:\horizon\exec}$ using different policy speed-up factors, and obtain the resulting observation $\observation_{\horizon\exec+1}$.
Notably, the pair $(\observation_{\horizon\exec+1}, \conditioningactions=\action_{\horizon\exec: \horizon\exec + \futureactionhorizon})$ represents the input to SAIL for CFG-based action generation. 
Instead of performing generation, we assess how well the action condition aligns with the observation.

To quantify this alignment, we follow the methodology described in \Cref{appendix:subsec:CFG_experiments}, where we sample $N=64$ action sequences from the unconditional action distribution.
We then compute in-distribution scores using standard out-of-distribution (OOD) techniques:
\begin{enumerate}
    \item \textbf{Kernel Density Estimation (KDE)}~\cite{scott2015multivariate}: 
    We estimate the likelihood of the action condition under the empirical distribution using a Gaussian kernel, with the bandwidth adaptively selected via Scott's rule.
    Higher values suggest better in-distribution.
    \item \textbf{k-Nearest Neighbors (kNN) Distance}~\cite{loftsgaarden1965nonparametric}: 
    We quantify how close the action condition is to its nearest neighbors in the dataset. 
    Specifically, it is computed as the average Euclidean distance to the $k=8$ nearest samples.
    Lower values suggest better in-distribution.
    \item \textbf{Maximum Mean Discrepancy (MMD)}~\cite{gretton2012kernel}: 
    We compute the discrepancy between the unconditional action distribution and the action-condition (modeled as a Dirac delta) using a Gaussian kernel with a bandwidth of 0.5.
    Lower values suggest better in-distribution.
\end{enumerate}

A key aspect of this experiment is that we reset the simulation before each rollout and evaluate only a single receding horizon step. 
This design isolates the direct effect of policy speed-up on action conditioning, avoiding confounding influences such as accumulated errors from prior rollouts. 
If the analysis were performed over an entire task execution, it would be difficult to disentangle whether action condition mismatches stem from speed-up itself or from historical execution deviations.

\begin{figure*}[htbp]
    \centering
    \includegraphics[width=0.9\textwidth]{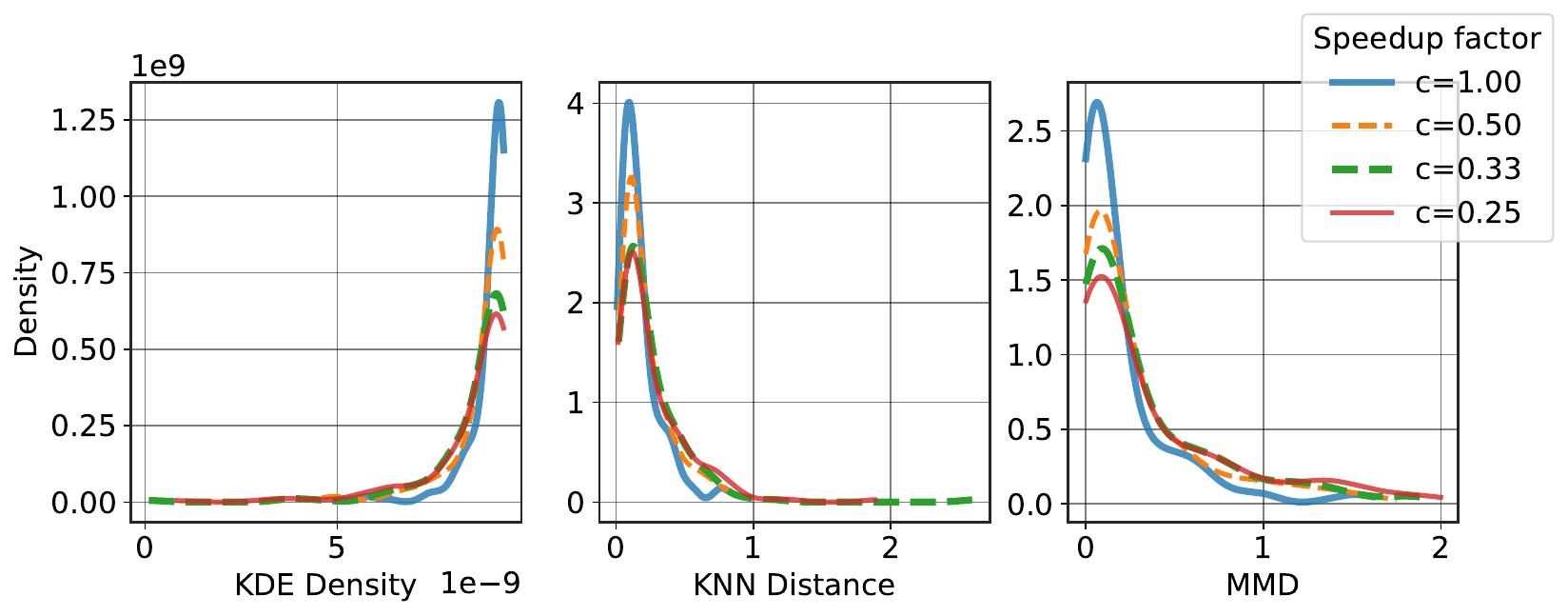} 
    \caption{
    \textbf{Effect of Policy Speed-up on Out-of-Distribution Action Conditions.}
    This figure evaluates how increasing the policy speed-up factor leads to mismatches between the action condition and the unconditional action distribution. 
    We assess this by computing how well action conditions derived from previously executed actions align with the unconditional action distribution given the current observation.
    We use three independent OOD detection metrics: Kernel Density Estimation (Left), k-Nearest Neighbor Distance (Middle), and Maximum Mean Discrepancy (Right).
    Across different metrics, we observe that as the policy speed-up factor increases, the action condition increasingly falls into the OOD region, exhibiting a long-tail distribution in high-OOD regions.
    Comparing normal-speed policies (blue) and highly accelerated policies (red) reveals that slower policies maintain better alignment between action conditions and current observations.
    }
    
    \label{fig:policy_speed_vs_ood}
\end{figure*}

We compute these metrics across 200 trials and visualize the density estimates in \Cref{fig:policy_speed_vs_ood}.
Our results confirm that as the policy speed-up factor increases, action conditions are more likely to fall outside the unconditional action distribution.
This trend suggests that at higher speeds, previously executed actions become less representative of the policy's expected next steps, leading to inconsistencies in action conditioning, and possible degradation of CFG performance.
Hence, we would require the need for adaptive mechanisms to mitigate conditioning mismatches in high-speed policy execution.

\subsection{Testing \hypothesis{H-EAG3}: Tracking error is highly correlated to observation-action misalignment}
\label{appendix:subsec:track_err}
Previous results indicate that action conditioning is most effective when the action condition is well-aligned with the current observation.
However, as policy speed-up increases, this alignment deteriorates, reducing the benefits of conditioning.
To effectively determine when action conditioning is beneficial, we need a reliable and efficient proxy for measuring misalignment.

A natural approach is to use the out-of-distribution (OOD) metrics introduced in \Cref{appendix:subsec:policy_speed_vs_fac}.
However, these methods are computationally expensive and impractical for real-time deployment.
Instead, we propose tracking error as a computationally efficient alternative.

\textbf{Defining Tracking Error.}
Given the current robot state $\state$ and the desired state indicated by the action $\action$, we define the position tracking error and orientation tracking error as follows:
\begin{align*}
    \trackerr_{pos} &= \norm{\action_{pos} - \state_{pos}} \\
    \trackerr_{ori} &= cos^{-1}(\frac{tr(R_{\triangle})-1}{2}), \quad R_{\triangle}=
    R(\action_{ori})\trans R(\state_{ori})    
\end{align*}
where $\state_{pos}$ and $\action_{pos}$ denote the real and desired end-effector positions, $\state_{ori}$ and $\action_{ori}$ represents the real and desired end-effector orientations.
$R(\cdot)$ maps any oreintation representation to an SO(3) rotation matrix, and $\text{tr}(\cdot)$ denotes the trace of a matrix.

\textbf{Analyzing Correlation with OOD Scores.} To assess whether tracking error serves as a reliable indicator of action condition misalignment, we compare tracking error values against the action condition's in-distribution scores computed using the metrics defined in \Cref{appendix:subsec:policy_speed_vs_fac}.
Specifically, we analyze the correlation separately for position and orientation tracking errors.

\begin{figure*}[htbp]
    \centering
    \includegraphics[width=0.9\textwidth]{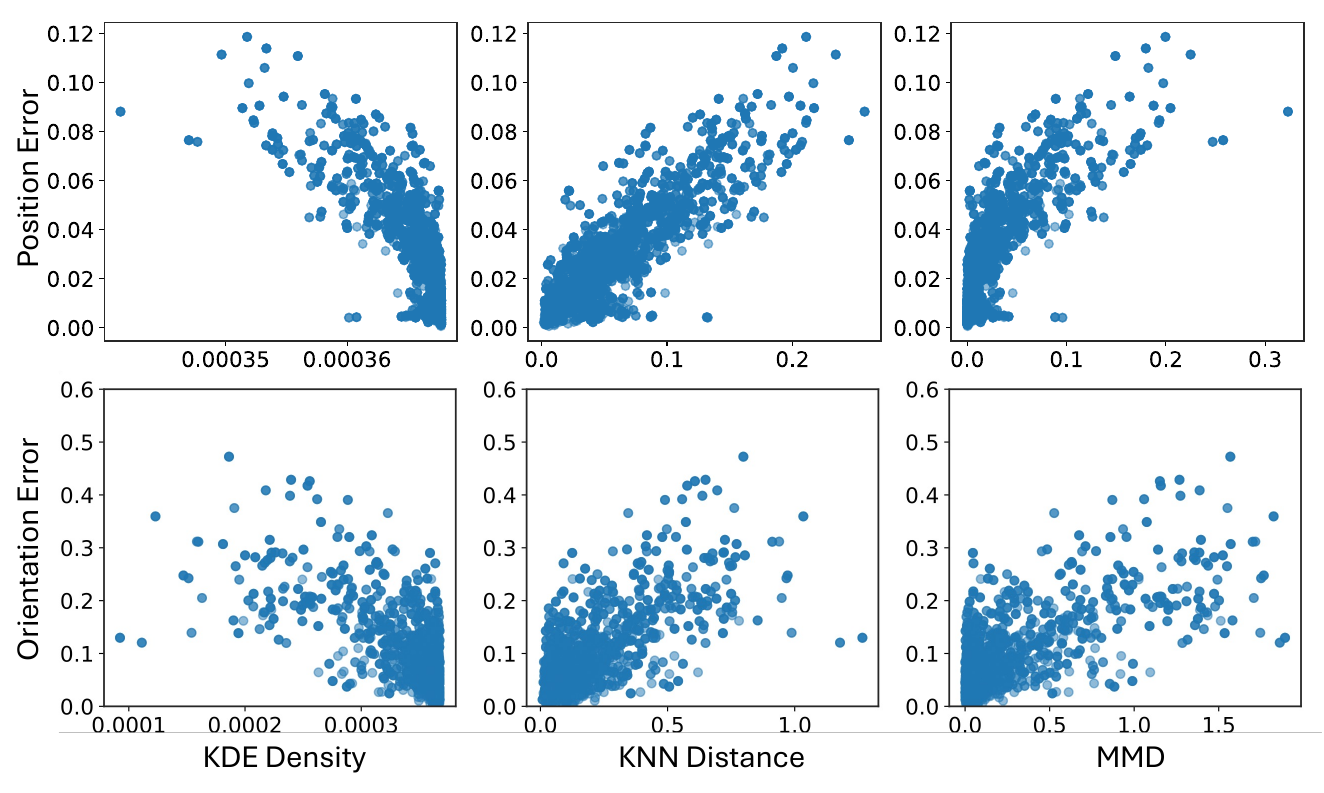} 
    \caption{
    \textbf{Correlation between Tracking Error and Out-of-Distribution Action Condition.}
        This figure illustrates the relationship between the tracking error and out-of-distribution (OOD) scores of action-conditions, computed using KDE Density (left), KNN distance (center), and MMD (right).
    The top row shows the correlation between the position tracking error and OOD scores in the position dimension.
    The bottom row shows the correlation between orientation tracking error and OOD scores in the orientation dimension.
    The result indicates that large tracking error corresponds to higher OOD scores, suggesting that tracking error can serve as a proxy for detecting misaligned action conditions.
    }
    \label{fig:track_err_vs_ood}
\end{figure*}

\Cref{fig:track_err_vs_ood} visualize these relationships.
The results indicate a clear trend: as tracking error increases, the action condition is more likely to fall outside the unconditional action distribution.
This suggests that tracking error is a useful proxy for detecting when action conditioning might degrade CFG performance.

\textbf{Impact of Adaptive CFG on Policy Performance.}
We evaluate whether adaptively applying CFG based on tracking error improves policy performance, particularly success rate and trajectory smoothness. 
To study the effect of different tracking error thresholds, we vary the position tracking error threshold across 0.01, 0.02, and 0.04 and evaluate the Square task with a policy speed-up factor of $c=0.33$. Each policy is tested over 50 scenarios, with three evaluations per scenario.

CFG is applied only when the tracking error is below the specified threshold, and we compare this approach to a baseline without guidance. 
Performance is measured by success rate and smoothness using the SPARC metric.
The proportion of inference steps where CFG was applied—referred to as the guided inference ratio—was 0.27, 0.47, and 0.78 for the three thresholds, respectively.

\begin{figure*}[htbp]
    \centering
    \includegraphics[width=0.9\textwidth]{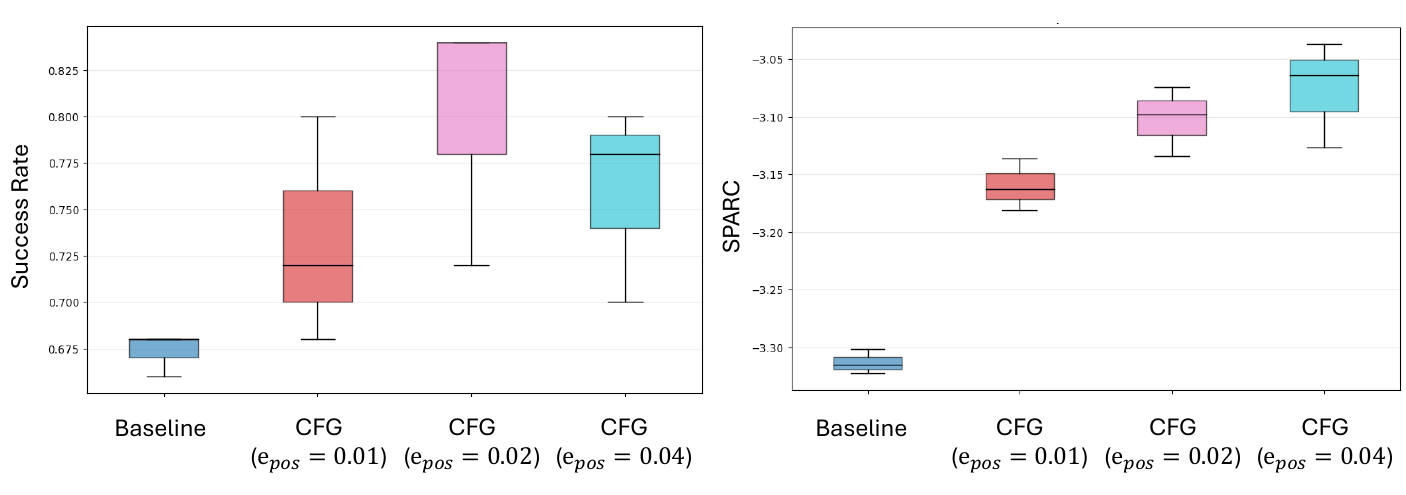} 
    \caption{
    \textbf{{Effect of Tracking Error Threshold on Policy Performance.}}
    This figure shows how different tracking error thresholds influence policy performance, measured by success rate (right) and trajectory smoothness (left).
    Each method is evaluated over 50 scenarios, repeated across three trials.
    The box plots display the maximum, mean, and minimum values per evaluation.
    The highest tracking error threshold (0.04) leads to a decline in success rate, while a more conservative threshold improves overall performance.
    }
    \label{fig:track_err_heuristic}
\end{figure*}

The results show that indiscriminate application of CFG (high guidance ratio) reduces success rate, likely due to misalignment between observation and action condition. 
Conversely, selectively applying CFG when the tracking error is low improves both success rate and smoothness. 
\Cref{fig:track_err_heuristic} illustrates that a moderate tracking error threshold leads to better overall performance, while higher thresholds degrade success rates.

One limitation is that the optimal tracking error threshold varies by task, as tasks with more complex reference trajectories naturally exhibit higher tracking errors. 
The values reported here were determined through hyperparameter tuning for the Square task. Nonetheless, these findings confirm that tracking error provides a useful heuristic for determining when to apply CFG, leading to more reliable policy execution.

\section{Aggregating actions}
\label{ssec:appendix:baseline}
We describe the algorithm for aggregating actions, which is one of our baselines, in \cref{alg:agg_alg}.

\begin{algorithm}[t]
  \caption{Aggregate Actions (baseline)}
  \label{alg:agg_alg}
  \DontPrintSemicolon              

  \KwIn{Sequence of delta Cartesian actions $A=\langle a_1,\dots,a_T\rangle$}

  $\textit{AggActions} \gets []$\;
  $\textit{curr} \gets a_1$\;

  \For{$a \in A[2{:}]$}{                     
      \If{$\bigl\lVert\textit{curr}\bigr\rVert > 0.05$ \text{cm}
           \textbf{or}
           $\operatorname{dot}(a,\textit{curr}) < 0.25$}{
          append $\textit{curr}$ to $\textit{AggActions}$\;
          $\textit{curr} \gets a$\;
      }\Else{
          $\textit{curr} \gets \textit{curr} + a$\;
      }
  }

  append $\textit{curr}$ to $\textit{AggActions}$\;
  \KwOut{Aggregated action sequence $\textit{AggActions}$}
\end{algorithm}

\section{Hyperparameters}
\label{appendix:hyperparameters}
The hyperparameters for our Policy backbone are listed in \Cref{tab:params}. The parameters for the consistency guiding is listed in \Cref{tab:consistency_params} and the controller parameters in simulation are listed in \Cref{tab:sim_controller_params}. The controller parameters for the real robot are listed in \Cref{tab:kp_kv_values}.

\begin{table}[t]
  \centering
  \caption{Key Parameters of Policy Architecture}
  \label{tab:params}
  \begin{tabular}{|l|c|}
    \hline
    \textbf{Parameter} & \textbf{Value} \\
    \hline
    \multicolumn{2}{|l|}{\textbf{General Settings}} \\
    \hline
    Algorithm & diffusion\_policy \\
    \hline
    Sequence Length & 32 \\
    \hline
    Frame Stack & 4 \\
    \hline
    Batch Size & 128 \\
    \hline
    Num. Epochs & 2000 \\
    \hline
    \multicolumn{2}{|l|}{\textbf{Horizon Settings}} \\
    \hline
    Observation Horizon & 4 \\
    \hline
    Action Horizon & 32 \\
    \hline
    Prediction Horizon & 32 \\
    \hline
    \multicolumn{2}{|l|}{\textbf{UNet \& Diffusion Settings}} \\
    \hline
    UNet Enabled & True \\
    \hline
    Diffusion Step Embed Dim & 256 \\
    \hline
    UNet Down-dimensions & 256, 512, 1024 \\
    \hline
    Kernel Size & 5 \\
    \hline
    \multicolumn{2}{|l|}{\textbf{EMA \& DDIM}} \\
    \hline
    EMA Enabled & True (power: 0.75) \\
    \hline
    DDIM Enabled & True \\
    \hline
    Train Timesteps & 100 \\
    \hline
    Inference Timesteps & 10 \\
    \hline
    Beta Schedule & squaredcos\_cap\_v2 \\
    \hline
    \multicolumn{2}{|l|}{\textbf{Future Action Conditioning}} \\
    \hline
    Enabled & True \\
    \hline
    Horizon & 4 \\
    \hline
    $p_{\text{cond}}$ & 0.3 \\
    \hline
    Weight & 1.0 \\
    \hline
    Null Token & zero \\
    \hline
    \multicolumn{2}{|l|}{\textbf{RGB Encoder Settings}} \\
    \hline
    Vision Encoder & ResNet18 \\
    \hline
    Pooling & kp = 32, temp = 1.0 \\
    \hline
    Randomizer & CropRandomizer (116$\times$116, 1 crop) \\
    \hline
  \end{tabular}
\end{table}\begin{table}[h]
    \centering
    \caption{Optimal Hyperparameter for Consistency Guiding}
    \label{tab:consistency_params}
    \renewcommand{\arraystretch}{1.2} 
    \begin{tabular}{|c|c|c|c|}
        \hline
        \textbf{Task} & \textbf{CFG weight} & \textbf{TEB (Ori.)} & \textbf{TEB (Pos.)} \\ 
        \hline
        Can   & 0 & 0.05   & 0.02   \\ 
        \hline
        Lift   & 1 & 0.05   & 0.02   \\ 
        \hline
        Square  & 1 & 0.05   & 0.02  \\ 
        \hline
        Mug Cleanup & 1 & 0.05  & 0.02  \\ 
        \hline
        Stack & 0 & 0.03  & 0.01  \\ 
        \hline
    \end{tabular}
\end{table}

\begin{table}[h]
\caption{Controller and SAIL parameters for each simulated task }
\label{tab:sim_controller_params}
\centering
\renewcommand{\arraystretch}{1.2} 
\begin{tabular}{|l|ccccc|}
\hline
                                  & \multicolumn{1}{l}{lift} & \multicolumn{1}{l}{can} & \multicolumn{1}{l}{square} & \multicolumn{1}{l}{stack} & \multicolumn{1}{l|}{mug cleanup} \\ \hline
$K_p$                                & 3000                     & 3000                    & 1000                       & 3000                      & 2000                            \\ \hline
damping                           & 0.5                      & 0.5                     & 1                          & .75                       & .75                             \\ \hline
\multicolumn{1}{|c|}{slowdown $c$} & 0.2                      & 0.5                     & 1.0                        & 0.2                       & 0.5                            \\ \hline
\end{tabular}
\end{table}

\begin{table}[]
    \centering
    \caption{Controller Gains for Demo Collection and SAIL Execution on Real Robot}
    \label{tab:kp_kv_values}
    \renewcommand{\arraystretch}{1.2} 
    \begin{tabular}{|l|c|c|}
        \hline
        \textbf{} & \textbf{Demo Collection} & \textbf{SAIL Execution} \\ \hline
        $K_p^{pos}$ & 150 & 300 \\ \hline
        $K_v^{pos}$ & 24.5 & 34.6 \\ \hline
        $K_p^{rot}$ & 250 & 400 \\ \hline
        $K_v^{rot}$ & 31.6 & 40.0 \\ \hline
    \end{tabular}
\end{table}

\begin{figure*}[htbp]
    \centering
    \begin{subfigure}[b]{0.4\textwidth}
        \centering
        \includegraphics[width=\textwidth]{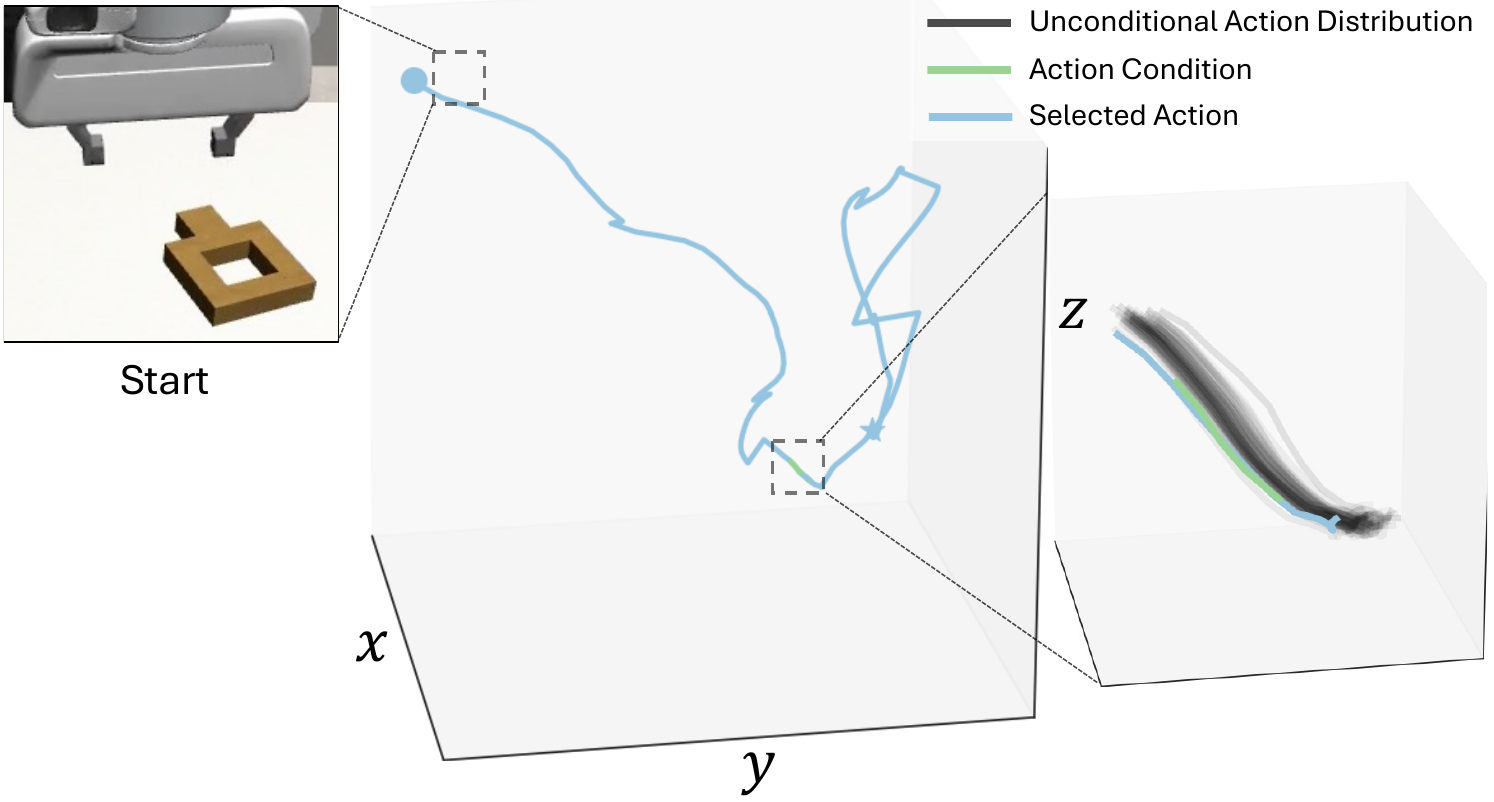}
        \caption{Successful rollout trajectory with action-conditioning.}
        \label{fig:cfg_qualitative_cfg}
    \end{subfigure}
    \vspace{1em}
    \begin{subfigure}[b]{0.4\textwidth}
        \centering
        \includegraphics[width=\textwidth]{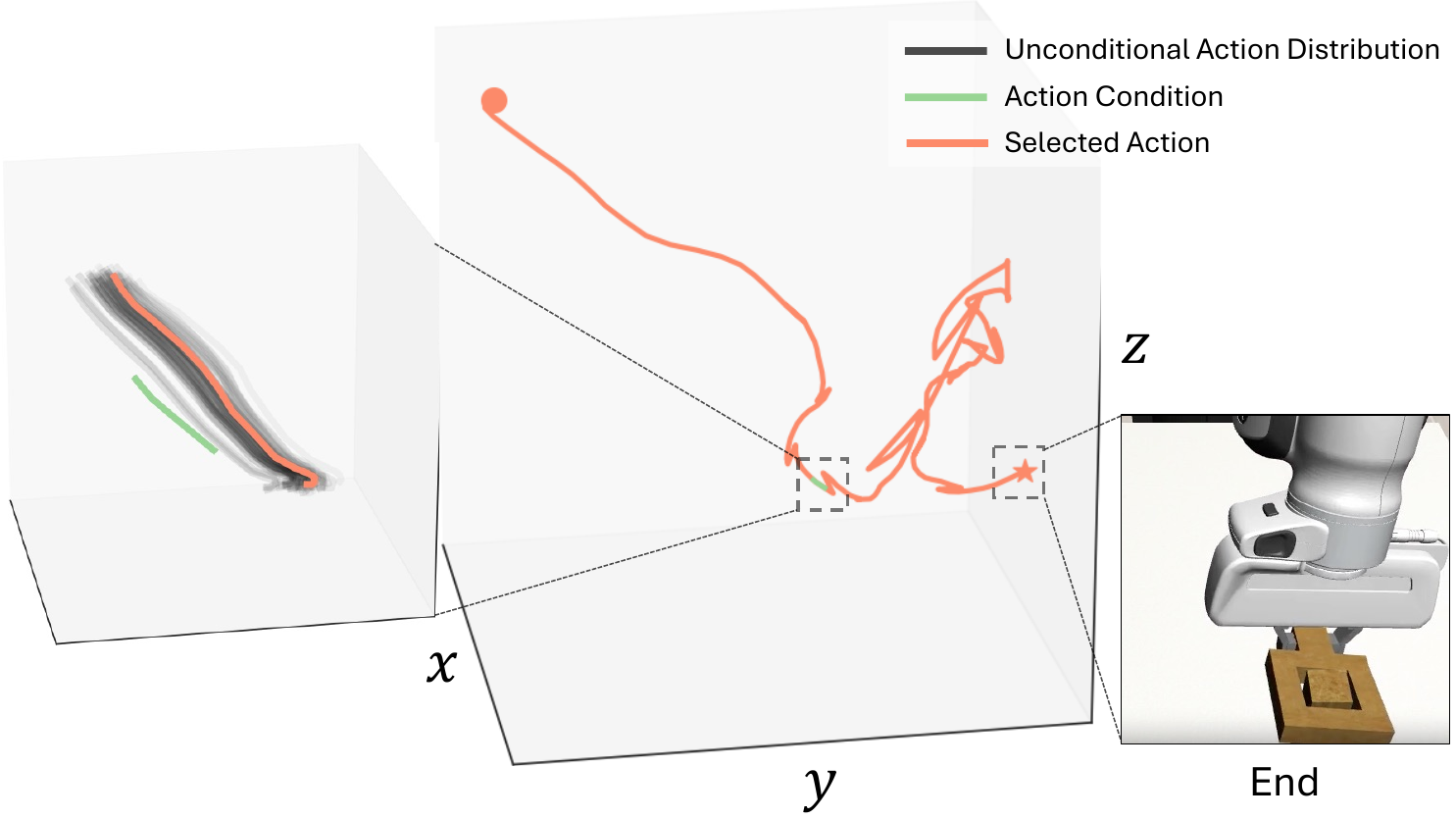}
        \caption{Successful rollout trajectory without action-conditioning.}
        \label{fig:cfg_qualitative_uncond}
    \end{subfigure}

    \caption{
        \textbf{Effect of Action-Conditioning on Smoothness of End-Effector Trajectories.}
        This figure illustrates end-effector trajectories, of a policy rollout on the square task, comparing scenarios with action-conditioning (\Cref{fig:cfg_qualitative_cfg}, blue trajectory) and without action-conditioning (\Cref{fig:cfg_qualitative_uncond}, red trajectory).
        Snapshots depicting the initial and final states of the square task are provided for each scenario.
        The guiding action (green line) and sampled unconditional actions (grey lines) are depicted, alongside the selected final action (solid colored lines).
        With action-conditioning, the chosen action closely aligns with the guided prediction (green), leading to smoother, goal-directed trajectories. Without action-conditioning, the final actions deviate significantly from the guide, resulting in less consistent trajectories.
        We further provide the snapshots of the initial state and the final state of the square task.
    }
    \label{fig:action_conditioning_effect}
\end{figure*}

\section{Detailed Ablation and SAIL with ACT}
\label{appendix:subsec:ablation}
A more detailed ablation of our method is provided in \Cref{tab:sim_ablation_full}. Furthermore, we present results of SAIL with ACT~\cite{zhao2023learning}, with the only difference being the use of temporal ensembling as opposed to CFG for maintaining consistency between consecutive predictions.
\begin{table}[ht]
\centering
\caption{Results of Ablation in Sim}
\label{tab:sim_ablation_full}
\begin{tblr}{
  colsep=3pt,
  width = \linewidth,
  colspec={c c *{6}{c}},
  cell{2}{1} = {r=4}{},
  cell{6}{1} = {r=4}{},
  cell{10}{1} = {r=4}{},
  cell{14}{1} = {r=4}{},
  cell{18}{1} = {r=4}{},
  vline{1,3,9} = {1-21}{},
  vline{3,9} = {3-5,7-9,11-13,15-17,19-21}{},
  hline{1-2,6,10,14,18,22} = {-}{},
}
       &     & \textbf{DP~\cite{chi2023diffusionpolicy}} & \textbf{SAIL} & \textbf{-HG} & \textbf{-AS} & \textbf{-C} & \textbf{Commanded Poses} \\
Lift   & SR $\uparrow$ & \textbf{1.00} & \textbf{1.00} & 0.67 & 0.97 & 0.98 & 0.89\\
       & TPR$\uparrow$  & 0.46 & \textbf{1.68} & 0.25 & {1.57} & 1.58 & 1.8\\
       & ATR$\downarrow$ & 2.23 & \textbf{0.61} & 3.39 &  {0.63} & 0.63 & 0.52\\
       & SOD$\uparrow$ & 1.08 & \textbf{3.98} & 0.71 &  {3.85} & 3.84 & 4.63\\
Can    & SR$\uparrow$  & \textbf{0.97} & 0.92 & 0.83 &   0.95 &  0.89 & 0.63\\
       & TPR$\uparrow$ & 0.18 & 0.51 & 0.18 &   \textbf{0.60} & 0.50 & 0.37\\
       & ATR$\downarrow$ & 5.52 & {1.81} & 4.36 &  \textbf{1.61} & 1.79 & 1.65\\
       & SOD$\uparrow$ & 1.05 & 3.20 & 1.33 &  \textbf{3.60} & 3.23 & 3.52\\
Square & SR$\uparrow$  & 0.83 & \textbf{0.86} & 0.59 &  {0.64} &  0.79 & 0.31\\
       & TPR$\uparrow$ & 0.10 & 0.13 & 0.06 &  \textbf{0.25} & 0.13 & 0.04\\
       & ATR$\downarrow$ & 7.56 & 6.41 & 8.8 & \textbf{2.50} & 5.78 & 4.25\\
       & SOD$\uparrow$ & 0.99 & 1.18 & 0.86 &  \textbf{3.01} &  1.31 & 1.77\\
Stack  & SR$\uparrow$  & \textbf{1.00} & 0.98 & 0.9 &   0.94 & 0.95 & 0.82\\
       & TPR$\uparrow$ & 0.19 & \textbf{0.66} & 0.15 &   0.61 & 0.62 & 0.43\\
       & ATR$\downarrow$ & 5.50 &  \textbf{1.56} & 6.6 &  1.71 & 1.56 & 2.81\\
       & SOD$\uparrow$ & 0.98 & \textbf{3.47} & 0.86 &   3.15 &  3.46 & 1.92\\
Mug    & SR$\uparrow$  & 0.68 &    \textbf{0.72} & 0.53 &   0.44 &  0.68 & 0.54\\
       & TPR$\uparrow$ & 0.03 &    \textbf{0.08} & 0.01 &  0.07 &  \textbf{0.08} & 0.03\\
       & ATR$\downarrow$ & 17.44 &     8.09 & 18.24 &  \textbf{5.37} &  8.15 & 17.38\\
       & SOD$\uparrow$ & 0.97 &        2.09 & 0.92 &    \textbf{3.14} & 2.07 & 0.92           
\end{tblr}
\end{table}


\begin{table}
\centering
\caption{ACT experiments with and without SAIL}
\label{tab:SAIL_ACT}

\begin{tblr}{
  width   = \linewidth,
  colspec = {|l|cccc|cccc|cccc|cccc|}, 
  rowsep  = 0.4pt,
  colsep  = 1pt,
  cell{1}{2}  = {c=4}{c},
  cell{1}{6}  = {c=4}{c},
  cell{1}{10} = {c=4}{c},
  cell{1}{14} = {c=4}{c},
  hline{1,3,5} = {-}{}
}

  &  Lift
  & & & &
     Can
  & & & &
     Square
  & & & &
     Stack \\

  & SR$\uparrow$ & TPR$\uparrow$ & ATR$\downarrow$ & SOD$\uparrow$
  & SR$\uparrow$ & TPR$\uparrow$ & ATR$\downarrow$ & SOD$\uparrow$
  & SR$\uparrow$ & TPR$\uparrow$ & ATR$\downarrow$ & SOD$\uparrow$
  & SR$\uparrow$ & TPR$\uparrow$ & ATR$\downarrow$ & SOD$\uparrow$ \\

  ACT  & \textbf{1.00} & 0.40 & 2.69 & 0.90
       & \textbf{0.77} & 0.12 & 4.04 & 0.96
       & \textbf{0.51} & 0.04 & 8.21 & 0.92
       & \textbf{0.73} & 0.11 & 6.18 & 0.87 \\

  SAIL & 0.94 & \textbf{1.50} & \textbf{0.65} & \textbf{3.71}
       & 0.56 & \textbf{0.25} & \textbf{2.12} & \textbf{2.72}
       & 0.43 & \textbf{0.11} & \textbf{3.22} & \textbf{2.34}
       & 0.68 & \textbf{0.28} & \textbf{3.14} & \textbf{1.72} \\

\end{tblr}
\end{table}

\end{document}